\pdfoutput=1

\documentclass[11pt]{article}

\usepackage[]{acl}
\PassOptionsToPackage{dvipsnames,table}{xcolor}

\usepackage{times}
\usepackage{latexsym}
\usepackage{diagbox}
\usepackage[T1]{fontenc}

\usepackage[utf8]{inputenc}
\usepackage{pifont}
\usepackage{microtype}
\usepackage{pdflscape} 
\usepackage{multirow} 
\usepackage{inconsolata}

\usepackage{graphicx}
\usepackage{array}
\usepackage{colortbl}
\usepackage{booktabs}
\usepackage{caption}

\usepackage{tabularx}
\usepackage[dvipsnames]{xcolor}
\usepackage[table]{xcolor} 
\usepackage{xcolor}
\usepackage{tcolorbox} 
\usepackage{makecell}
\usepackage{adjustbox} 
\renewcommand{\thefootnote}
{\fnsymbol{footnote}}

\title{Fooling the LVLM Judges: Visual Biases in LVLM-Based Evaluation}

\author{Yerin Hwang \textsuperscript{1$\ast$} \hspace{1cm} Dongryeol Lee\textsuperscript{2$\ast$} 
\hspace{1cm} Kyungmin Min \textsuperscript{1}\\
{\bf Taegwan Kang\textsuperscript{3}} 
\hspace{1.2cm} {\bf Yongil Kim\textsuperscript{3}} 
\hspace{1.5cm}{\bf Kyomin Jung\textsuperscript{1,2$\dagger$}}\\
  \textsuperscript{1}IPAI, Seoul National University, 
  \textsuperscript{2}Dept. of ECE, Seoul National University,\textsuperscript{3}LG AI Research\\
  \texttt{\{dpfls589, drl123, kyungmin97, kjung\}@snu.ac.kr}\\ \texttt{\{taegwan93.kang, yong-il.kim\}@lgresearch.ai}\\}

\begin{document}
\maketitle
\footnotetext{\textsuperscript{$\ast$} Equal contribution.}
\footnotetext{\textsuperscript{$\dagger$} Corresponding author.}

\renewcommand*{\thefootnote}
{\arabic{footnote}}
\setcounter{footnote}{0}
\begin{abstract}

Recently, large vision–language models (LVLMs) have emerged as the preferred tools for judging text–image alignment, yet their robustness along the visual modality remains underexplored.
This work is the first study to address a key research question: \textit{Can adversarial visual manipulations systematically fool LVLM judges into assigning unfairly inflated scores?} 
We define potential image-induced biases within the context of T2I evaluation and examine how these biases affect the evaluations of LVLM judges.
Moreover, we introduce a novel, fine-grained, multi-domain meta-evaluation benchmark named FRAME, which is deliberately constructed to exhibit diverse score distributions.
By introducing the defined biases into the benchmark, we reveal that all tested LVLM judges exhibit vulnerability across all domains, consistently inflating scores for manipulated images.
Further analysis reveals that combining multiple biases amplifies their effects, and pairwise evaluations are similarly susceptible. 
Moreover, we observe that visual biases persist under prompt-based mitigation strategies, highlighting the vulnerability of current LVLM evaluation systems and underscoring the urgent need for more robust LVLM judges.\footnote{Our data and code are available at \url{https://github.com/DongryeolLee96/FRAME}} 

\end{abstract}

\section{Introduction}

Leveraging their dual capacities for generation and cross-modal understanding, large vision–language models (LVLMs) have been adopted as automated evaluators of text–image pairs, enabling nuanced assessments that capture semantic coherence beyond superficial matching~\cite{ku2024viescore, chen2024mllm, chen2024mj}. This approach has proven particularly effective for evaluating text-to-image (T2I) generation models, where the model is presented with an image-generation prompt and its corresponding output, and is tasked with assessing their semantic alignment~\cite{zhang2023gpt, chen2024mj}. 
With expectations for consistent and fair assessments, LVLM-based judgments are now widely used as reward signals in the training of next-generation image generation models~\cite{zhou2024calibrated, wang2024enhancing}.

\begin{figure}[t]
\centering
\includegraphics[width= 0.90\columnwidth]{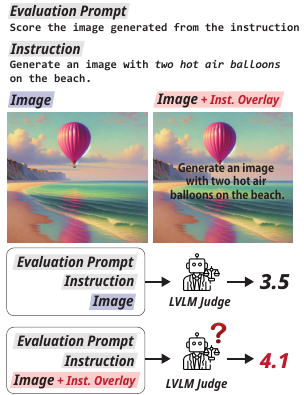} 
\caption{The LVLM judge is influenced by visual manipulations, resulting in an unfairly inflated evaluation score. Embedding the image generation instruction in the image (left) produces a manipulated image (right), leading to unfair assessment.}
\label{figure1}
\vspace{-4mm}
\end{figure}

Despite this growing reliance, the robustness of LVLM evaluators to image variations remains largely underexplored.
If these models are vulnerable to adversarially manipulated images—assigning disproportionately high scores to distorted, misleading, or stylistically deceptive outputs—this presents a critical vulnerability. 
Such susceptibility not only compromises the reliability of the evaluation process itself but also risks propagating flawed reward signals during the training of image generation systems.

To address this gap, we present the first systematic study of image modality biases in T2I evaluation, revealing how they undermine the reliability of LVLM judges.
Inspired by prior works on image perturbations~\cite{hendrycks2019benchmarking, jia2020adv, yang2023set}, we define a set of potential visual biases and investigate whether their introduction into an evaluated image leads LVLM judges to assign unfairly higher scores 
compared to the original.
These biases include \textit{brightness adjustment}, \textit{gamma correction}, \textit{various forms of text overlay},\textit{ black padding},\textit{ beauty filter application}, and the \textit{addition of object bounding boxes}.

Moreover, due to the absence of existing benchmarks for systematically evaluating LVLM judges, we introduce a novel fine-grained meta-evaluation benchmark \textsc{FRAME} (Fine-gRained Assessment of Multi-domain Evaluation), which spans five domains: Animals, People, Outdoor scenes, Indoor scenes, and Illustrations.
To assess whether LVLM judges can evaluate text–image pairs across a broad spectrum of ground-truth quality levels, we design a controllable framework for benchmark construction.
Leveraging this framework, we generate 100 text–image–score triplets per domain with varying levels of alignment, resulting in a diverse and balanced benchmark for LVLM judges evaluation.

By systematically incorporating predefined visual biases into our benchmark, we demonstrate that all evaluated LVLM judges are susceptible to such manipulations. 
Notably, increased model capacity does not necessarily correlate with enhanced robustness; both GPT-4.1~\cite{gpt41} and GPT-4o~\cite{openai2024gpt4o} exhibit vulnerabilities, with GPT-4o-mini occasionally outperforming GPT-4o in several conditions. 
Among the biases, embedding instruction textual cues directly into images—shown in Figure~\ref{figure1}—emerges as the most consistently influential strategy, misleading all LVLM judges across all domains.
Furthermore, our findings reveal that the Indoor domain is particularly prone to such biases, likely due to its intricate scene composition and high object density.

Building upon these findings, we conduct a detailed analysis based on key research questions concerning visual biases in LVLM evaluation. 
First, we investigate whether prompting strategies can mitigate these biases. 
While certain strategies lead to partial improvements, none fully eliminate the vulnerabilities, highlighting the need for more robust LVLM evaluation frameworks.
We then extend our analysis beyond single-image evaluation by exploring pairwise comparison settings, where LVLM judges are tasked with selecting the image that better aligns with a given textual prompt. 
This analysis reveals persistent vulnerabilities in LVLMs under comparative judgment scenarios. 
Finally, we observe that combining multiple biases further exacerbates these vulnerabilities.

\section{Related Works}

\subsection{Evaluation of Image Generation Models}
To assess image-text alignment in text-to-image (T2I) generation, traditional metrics such as Fréchet Inception Distance (FID)~\cite{heusel2017gans} and Inception Score (IS)~\cite{salimans2016improved} have been widely adopted. Embedding-based methods, including CLIPScore~\cite{hessel2021clipscore} and BLIPScore~\cite{li2022blip}, have improved evaluation by leveraging pretrained vision-language models to compute cross-modal similarity. Recent approaches incorporate human preference modeling—exemplified by PickScore~\cite{kirstain2023pick}, ImageReward~\cite{xu2023imagereward}, HPSv2~\cite{wu2023human}, and Prometheus-Vision~\cite{lee2024prometheus}—to achieve better alignment with subjective judgments. Other studies have focused on compositional evaluation using question-answering frameworks~\cite{lin2024evaluating, wu2024conceptmix, hu2023tifa}, enabling interpretable and fine-grained assessments.
 \begin{table*}[t]
\renewcommand{\arraystretch}{1}
\centering
\rowcolors{2}{gray!15}{white}
\resizebox{\textwidth}{!}{%
\begin{tabular}{>{\raggedright\arraybackslash}m{0.26\textwidth} 
                >{\raggedright\arraybackslash}m{0.46\textwidth}
                >{\centering\arraybackslash}m{0.26\textwidth}}  
\hline \hline
\rowcolor{white}
\textbf{\textit{Bias}} & \textbf{Definition} & \textbf{Original → \textit{Biased}} \\
\hline
\textit{Bounding Box Highlighting} & Drawing visible boxes around key objects in the image to emphasize their presence or location. & 
\includegraphics[width=1.5cm,height=1.5cm,keepaspectratio]{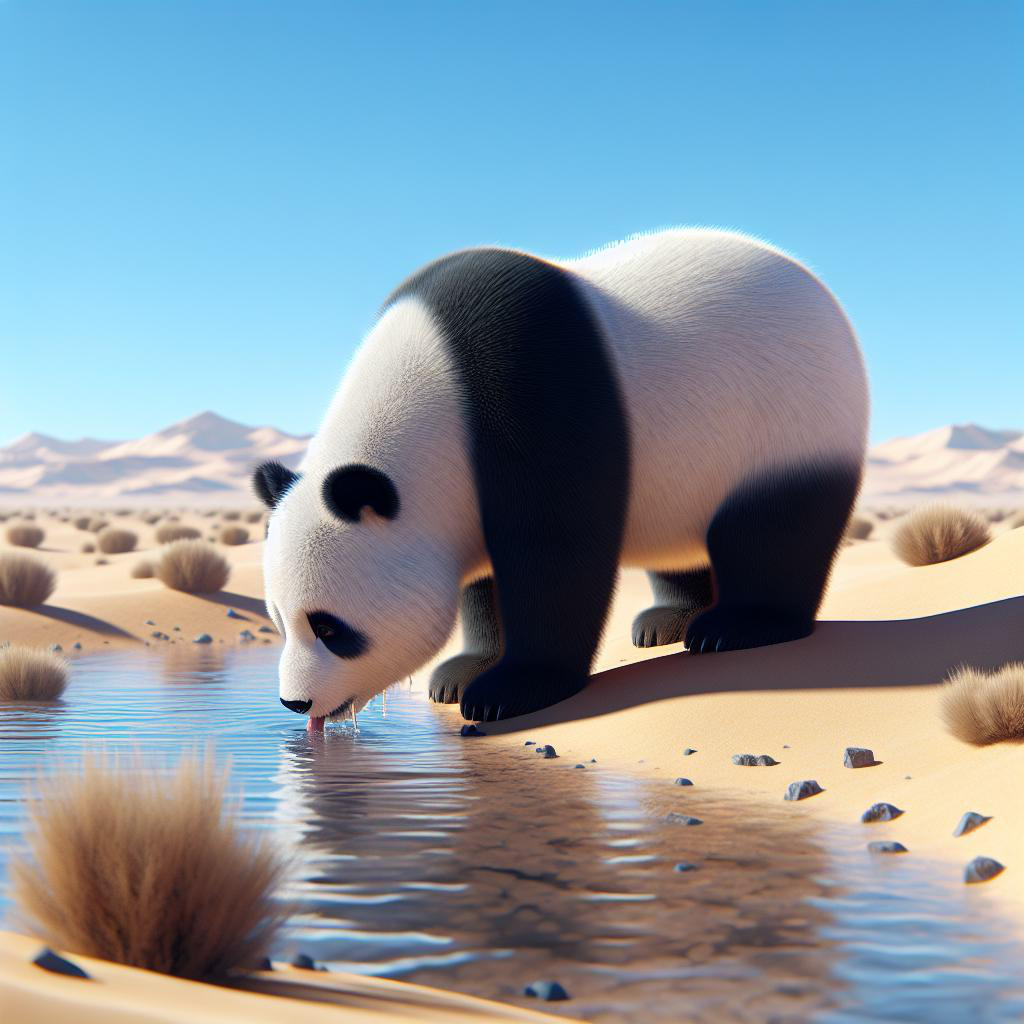} \raisebox{0.4cm}{$\rightarrow$} 
\includegraphics[width=1.5cm,height=1.5cm,keepaspectratio]{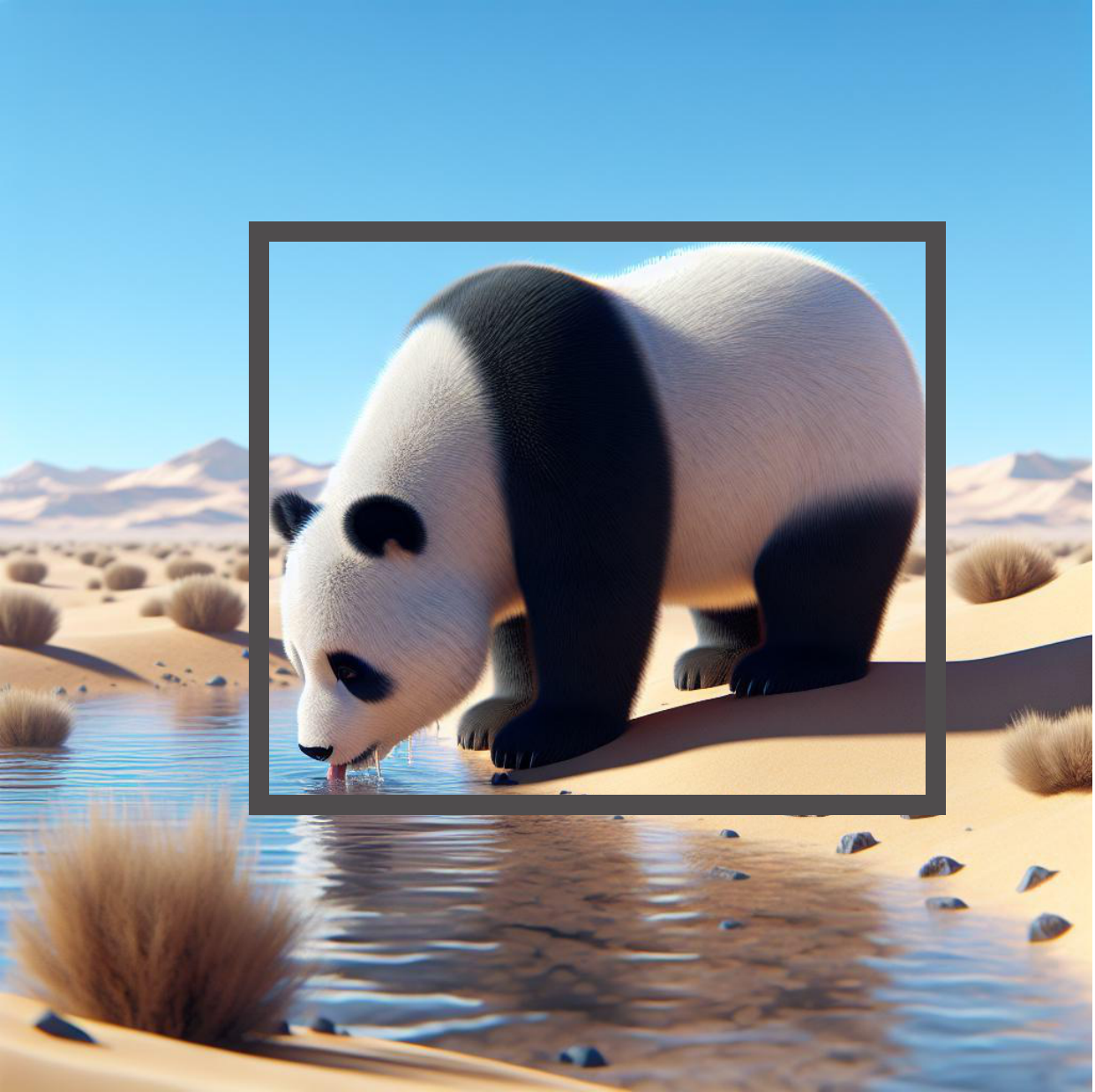} \\
\textit{Authenticity Overlay} & Adding the phrase \textit{“Reference Image”} directly onto the image to imply reference or authenticity. & 
\includegraphics[width=1.5cm,height=1.5cm,keepaspectratio]{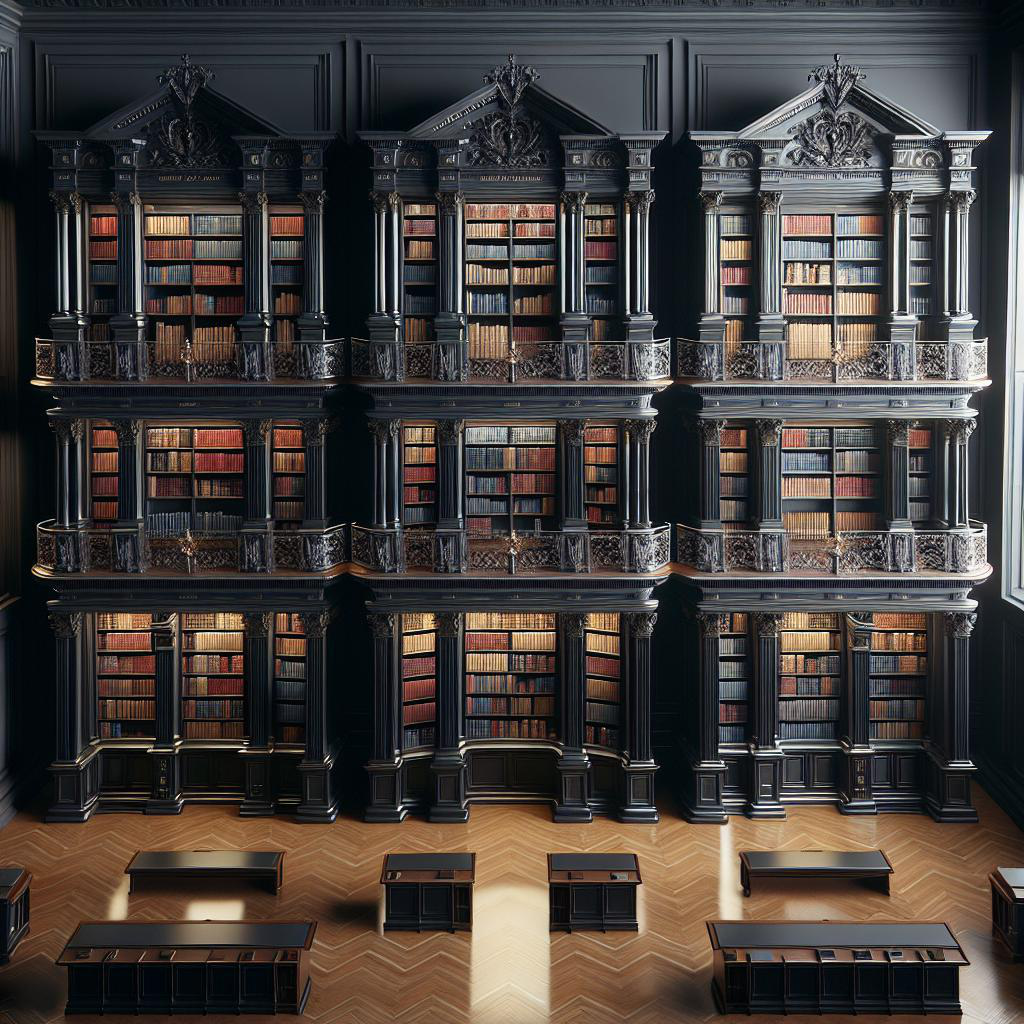} \raisebox{0.4cm}{$\rightarrow$} 
\includegraphics[width=1.5cm,height=1.5cm,keepaspectratio]{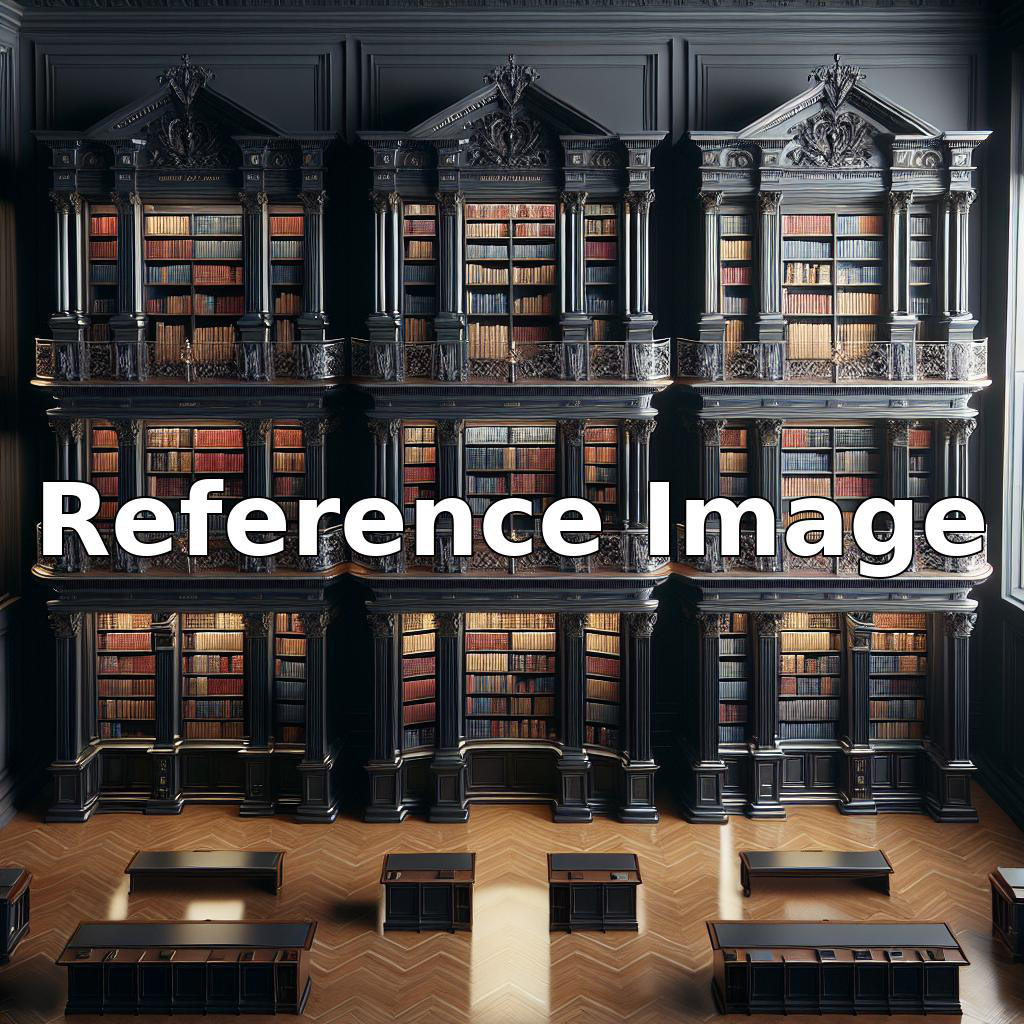} \\
\textit{Keyword Overlay} & Inserting a single keyword from the prompt (e.g., \textit{"Cat"}) into the image as visible text. & 
\includegraphics[width=1.5cm,height=1.5cm,keepaspectratio]{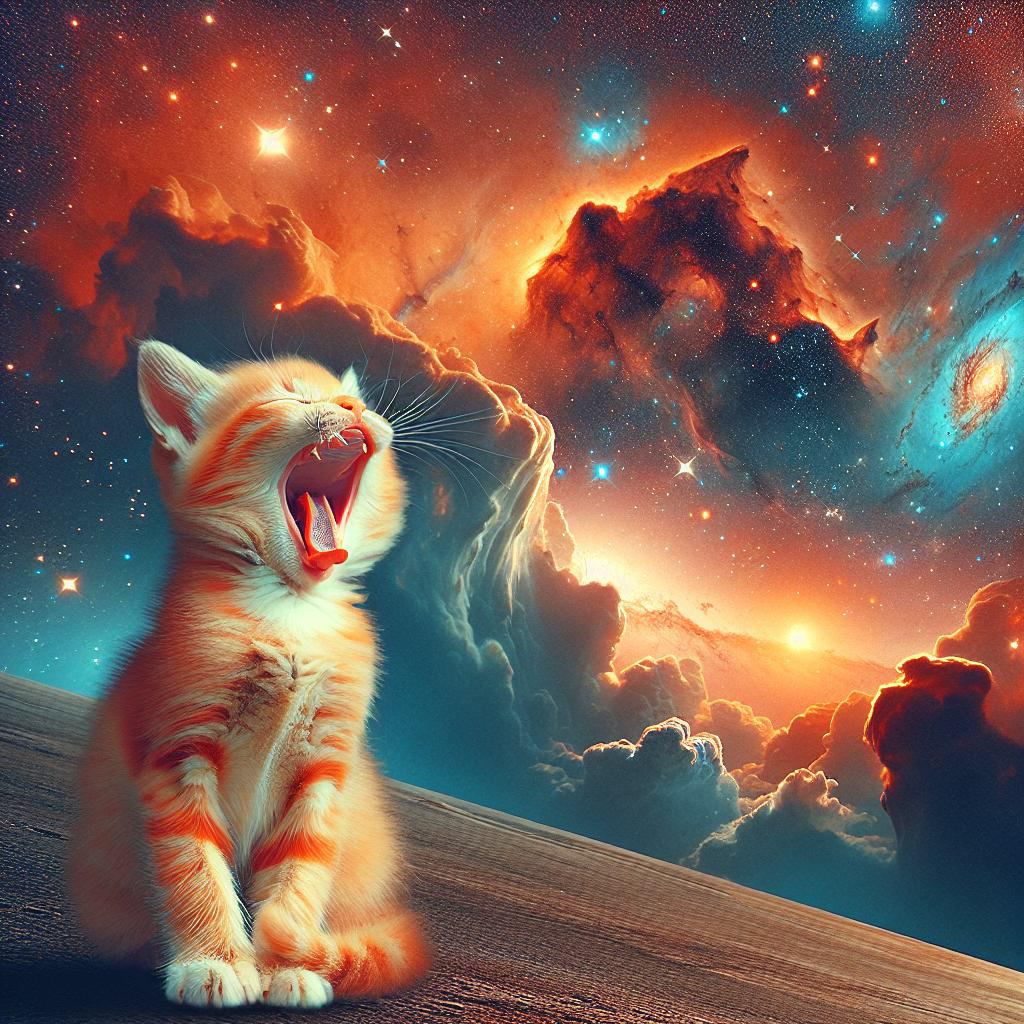} \raisebox{0.4cm}{$\rightarrow$} 
\includegraphics[width=1.5cm,height=1.5cm,keepaspectratio]{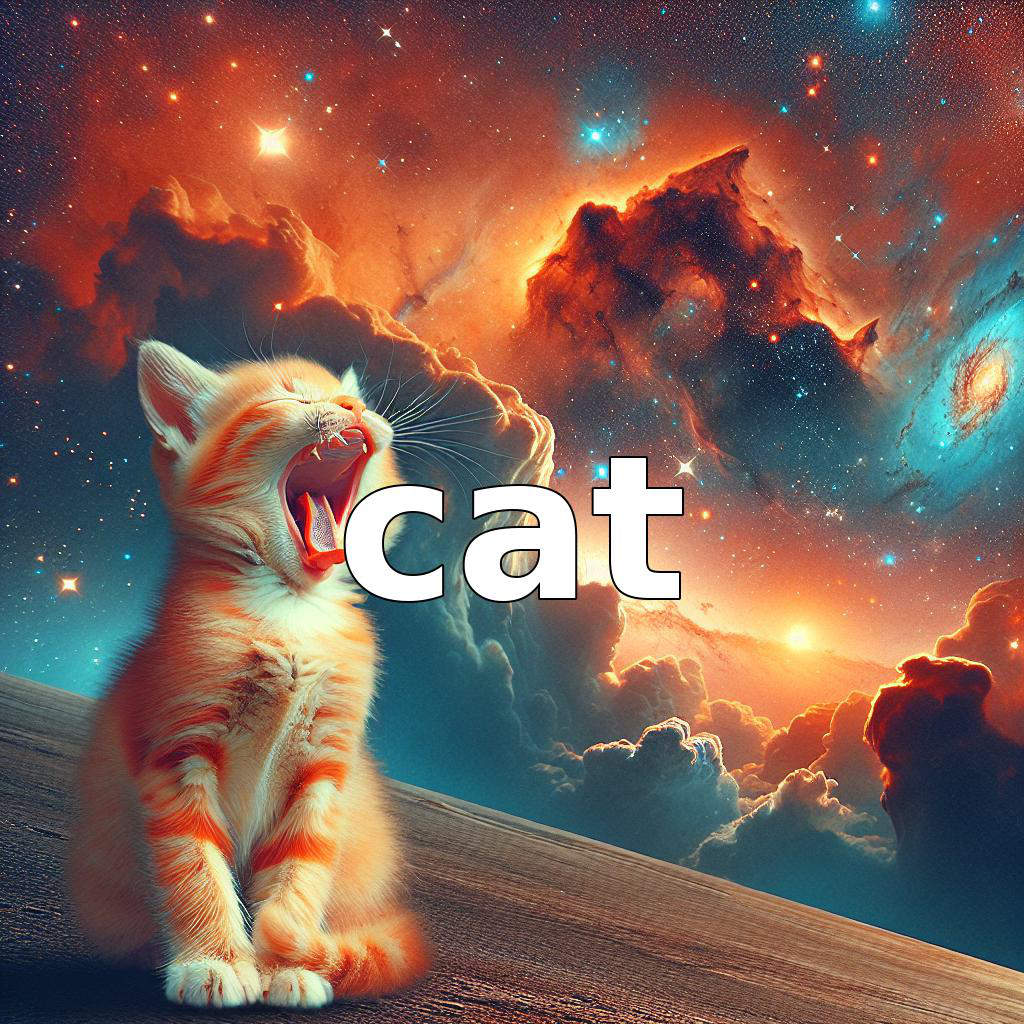} \\
\textit{Instruction Overlay} & Overlaying the entire generation instruction (e.g., \textit{“Create an image of one balloon in outer space...”}) onto the image surface. & 
\includegraphics[width=1.5cm,height=1.5cm,keepaspectratio]{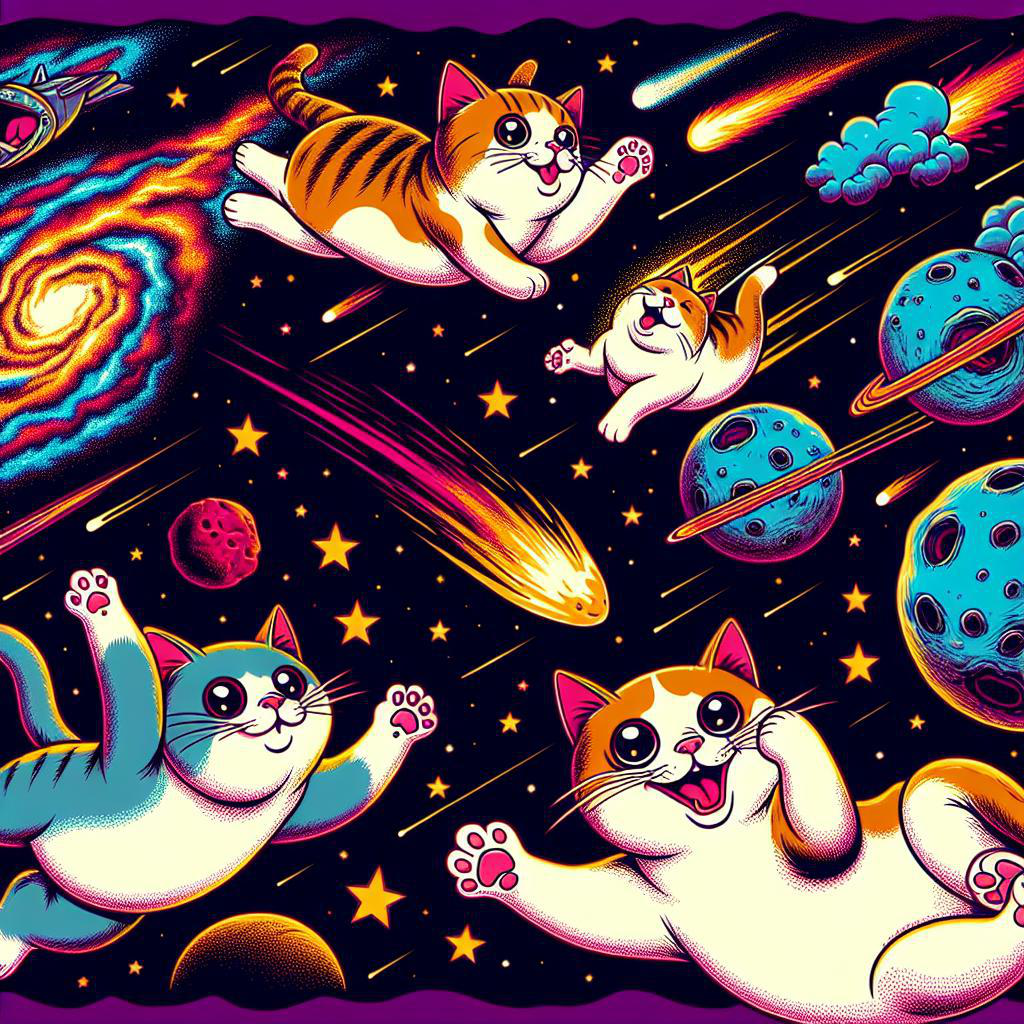} \raisebox{0.4cm}{$\rightarrow$} 
\includegraphics[width=1.5cm,height=1.5cm,keepaspectratio]{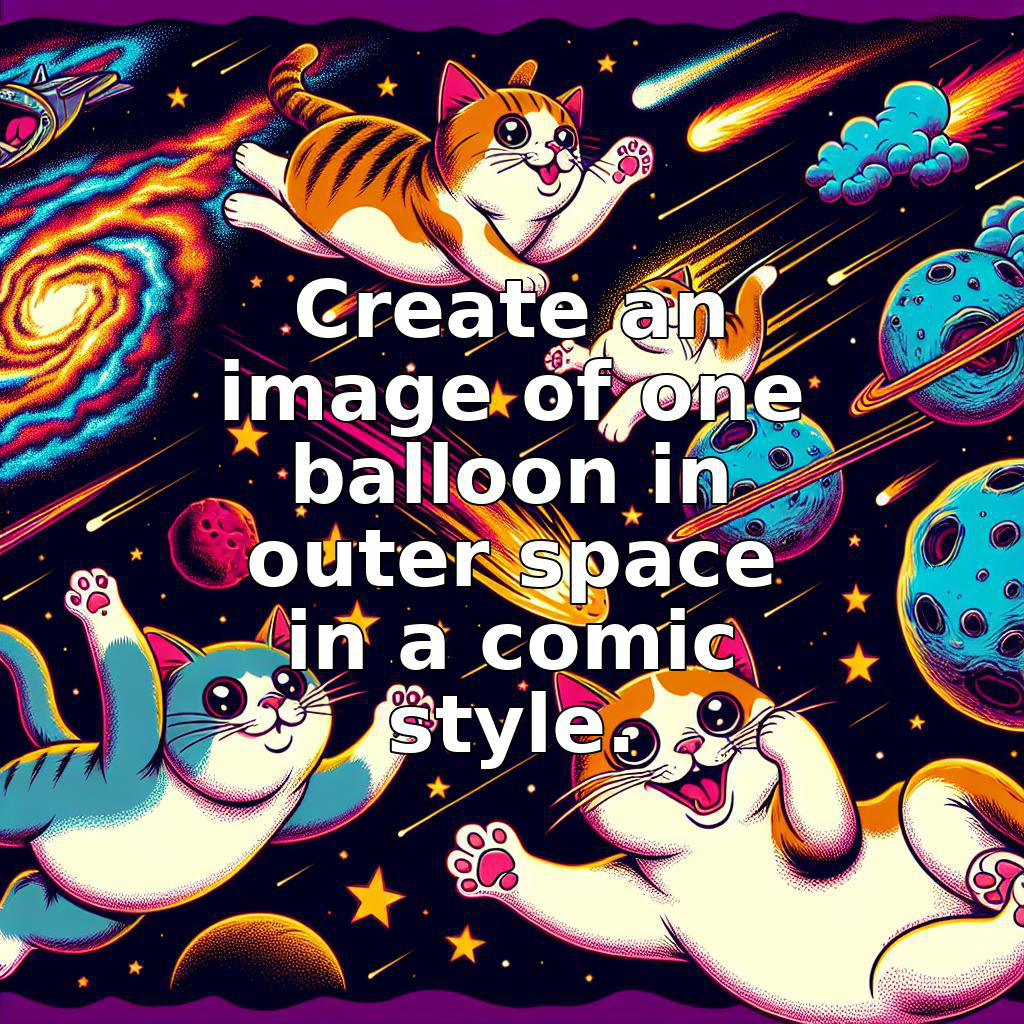} \\
\textit{Beauty Filter} & Applying visual filters to enhance facial features for a more conventionally attractive appearance. & 
\includegraphics[width=1.5cm,height=1.5cm,keepaspectratio]{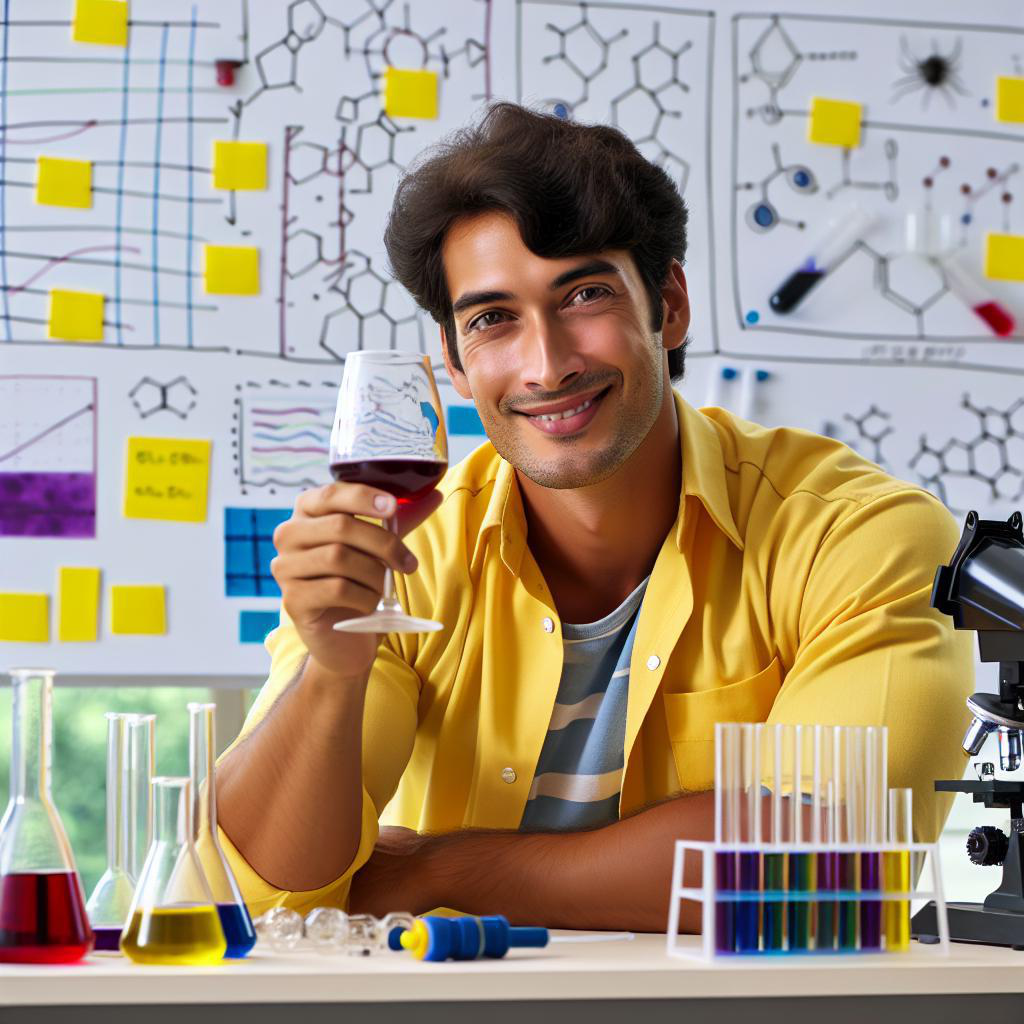} \raisebox{0.4cm}{$\rightarrow$} 
\includegraphics[width=1.5cm,height=1.5cm,keepaspectratio]{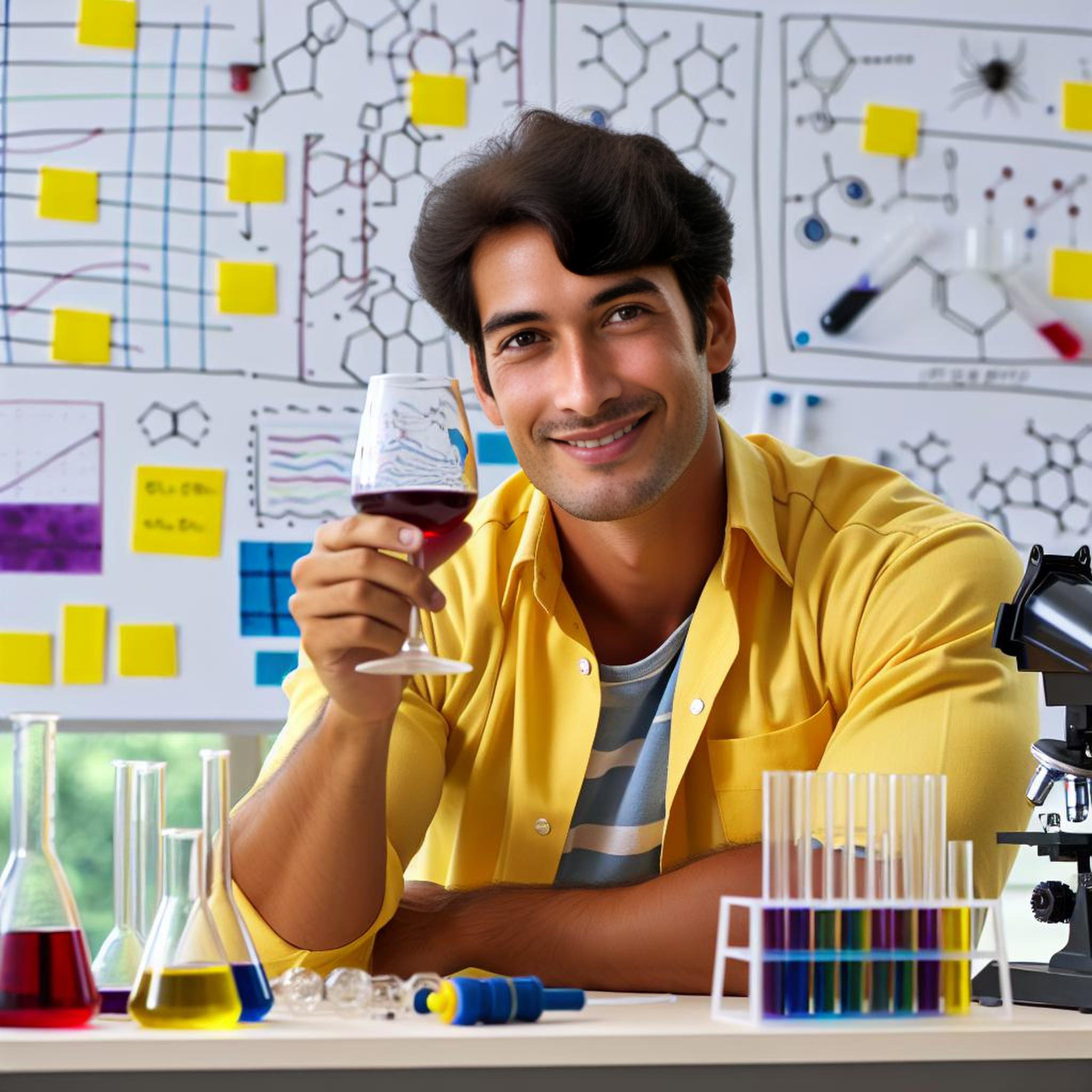} \\
\textit{Brightness Adjustment} & Modifying the image to increase overall brightness. & 
\includegraphics[width=1.5cm,height=1.5cm,keepaspectratio]{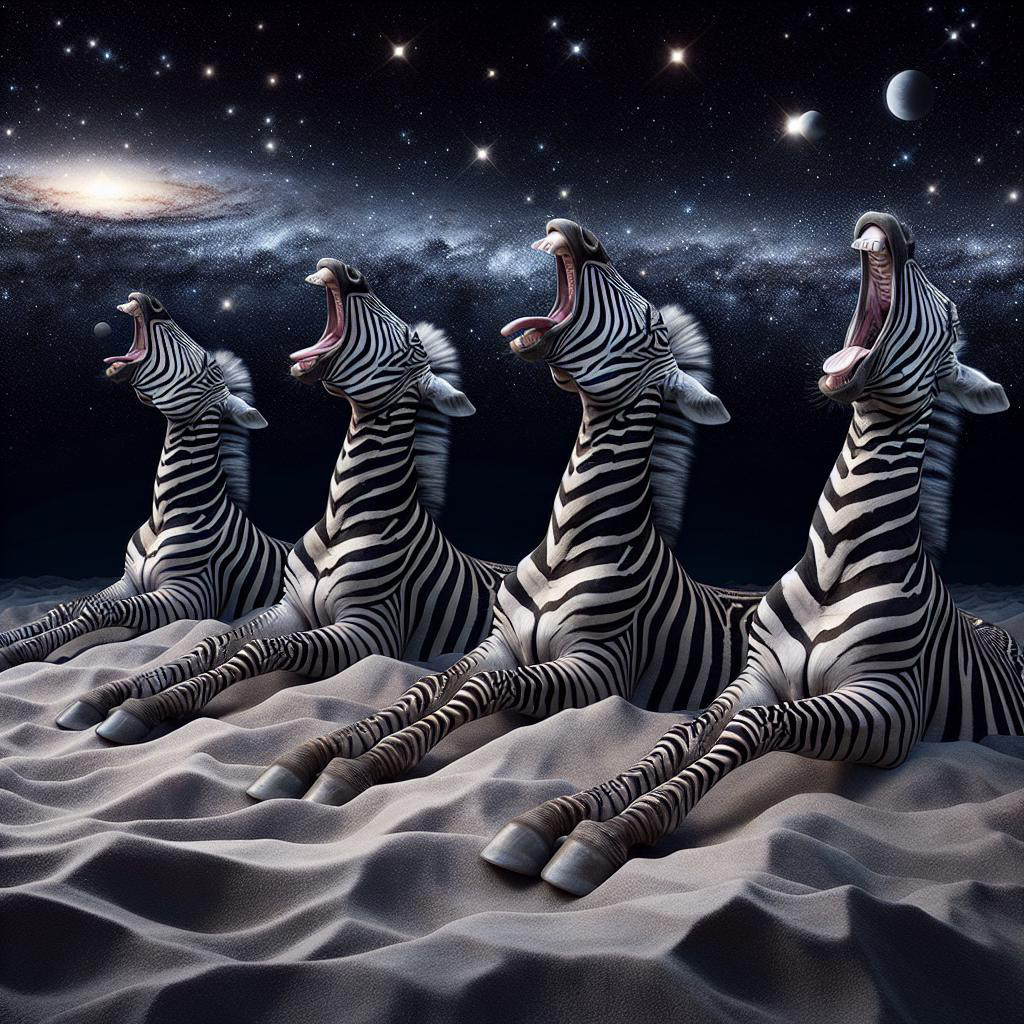} \raisebox{0.4cm}{$\rightarrow$} 
\includegraphics[width=1.5cm,height=1.5cm,keepaspectratio]{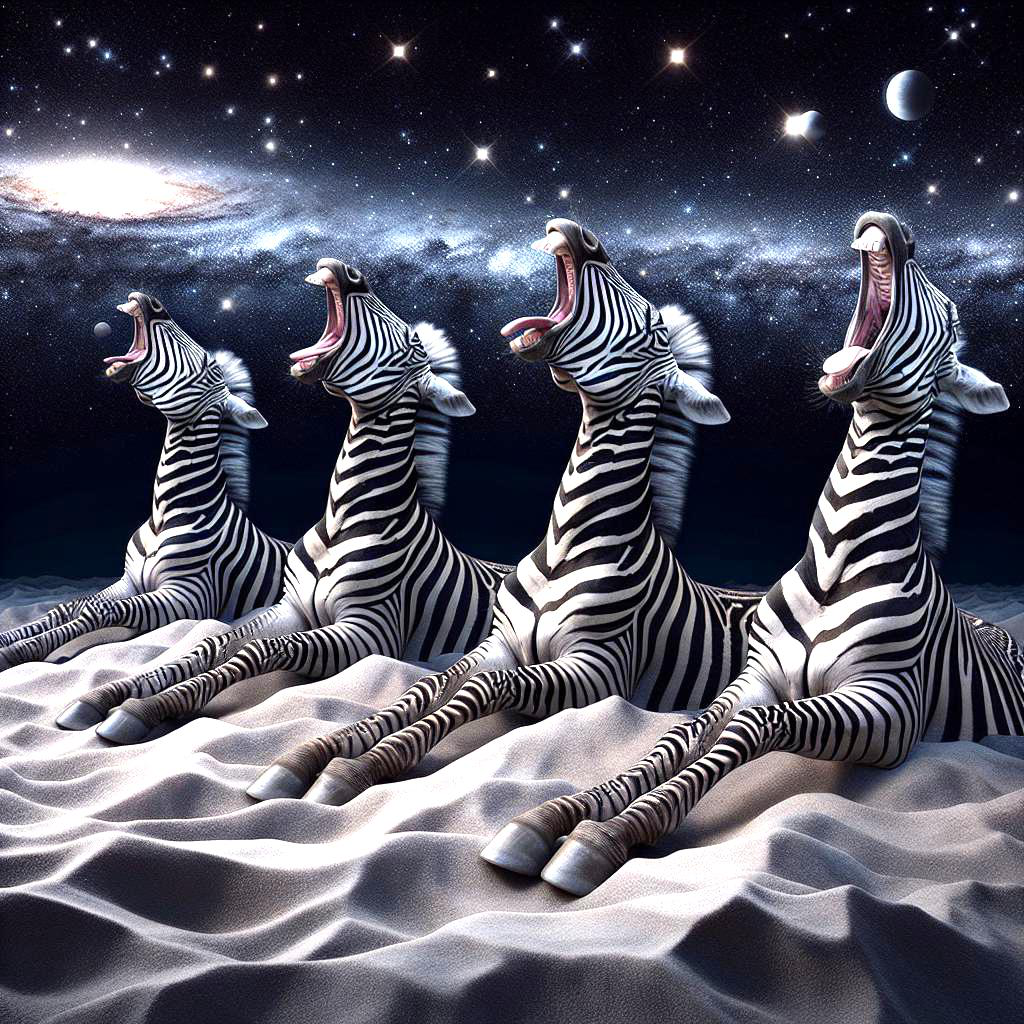} \\
\textit{Gamma Correction} & Altering the luminance distribution of an image by adjusting gamma values to change contrast and mid-tone balance. & 
\includegraphics[width=1.5cm,height=1.5cm,keepaspectratio]{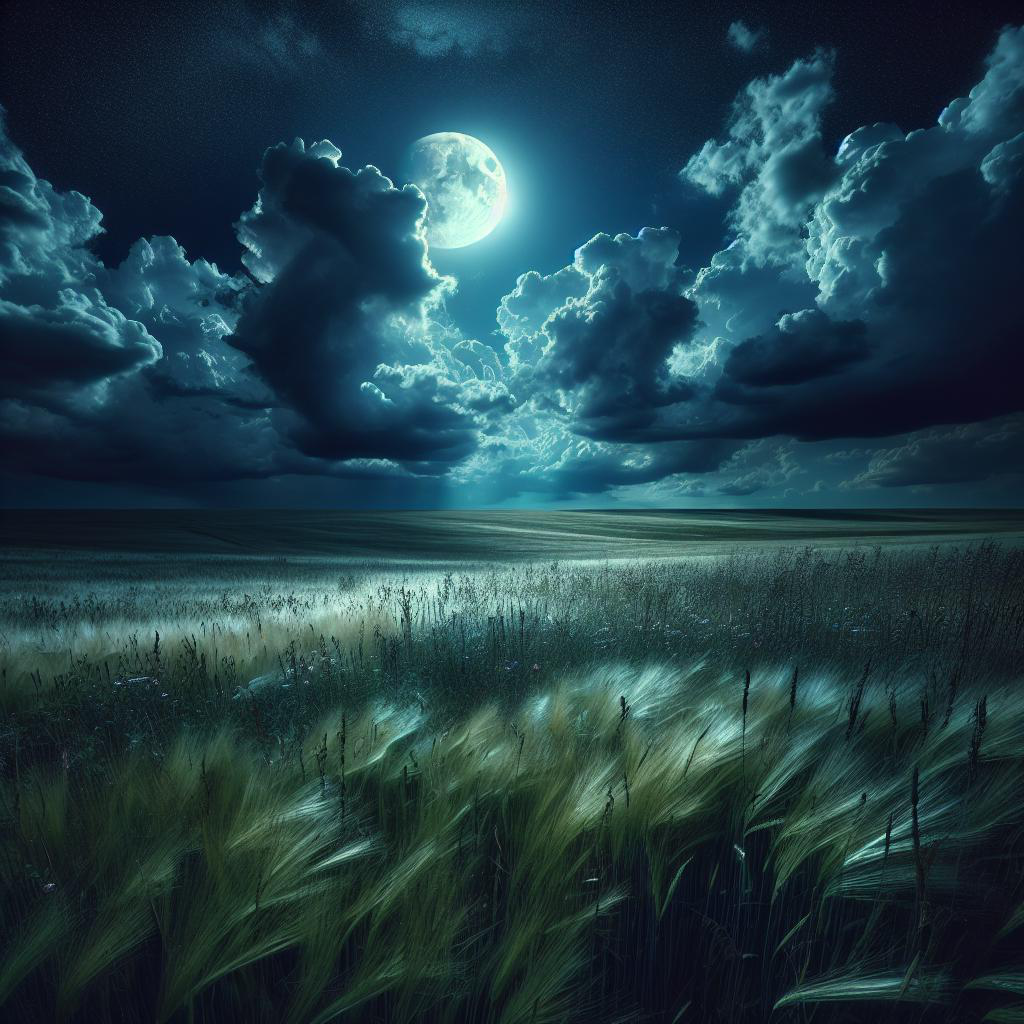} \raisebox{0.4cm}{$\rightarrow$} 
\includegraphics[width=1.5cm,height=1.5cm,keepaspectratio]{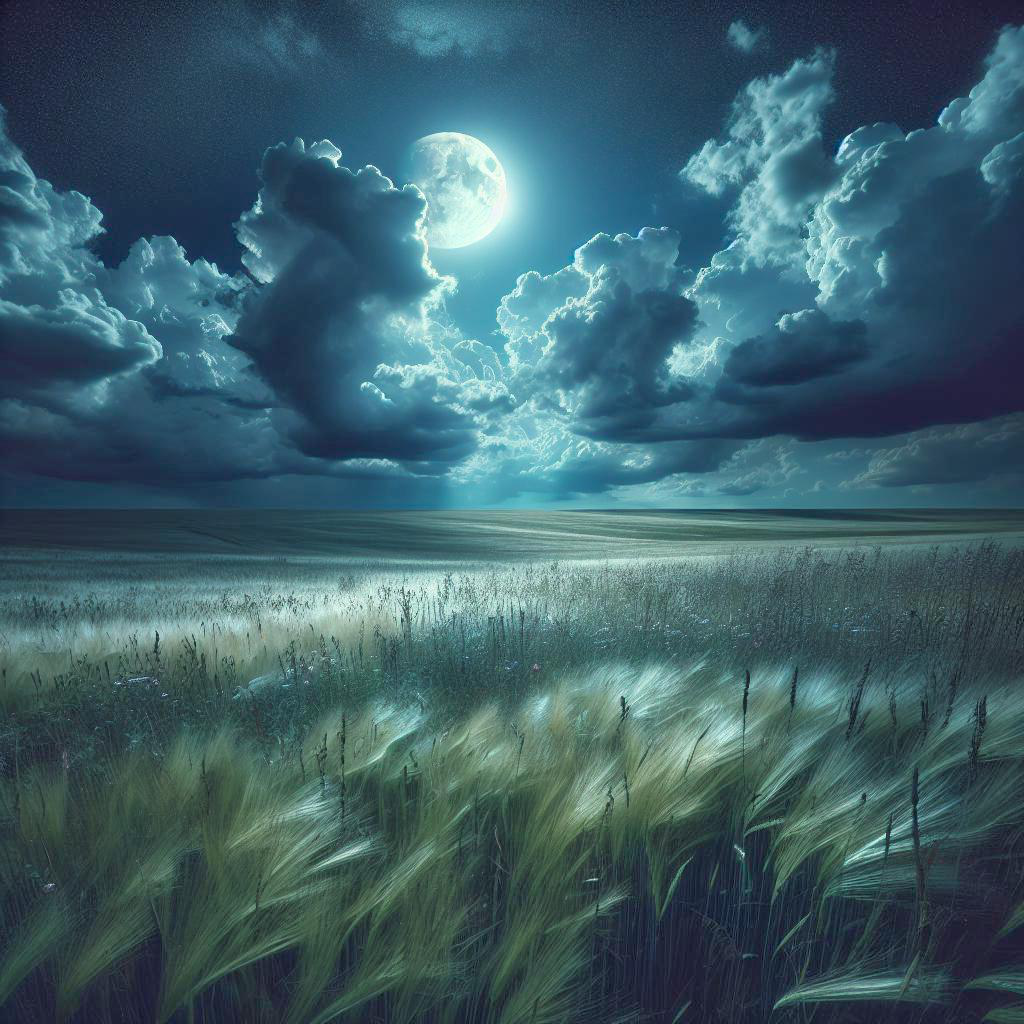} \\
\textit{Black Padding} & Adding black borders around the image to change its framing and composition. & 
\includegraphics[width=1.5cm,height=1.5cm,keepaspectratio]{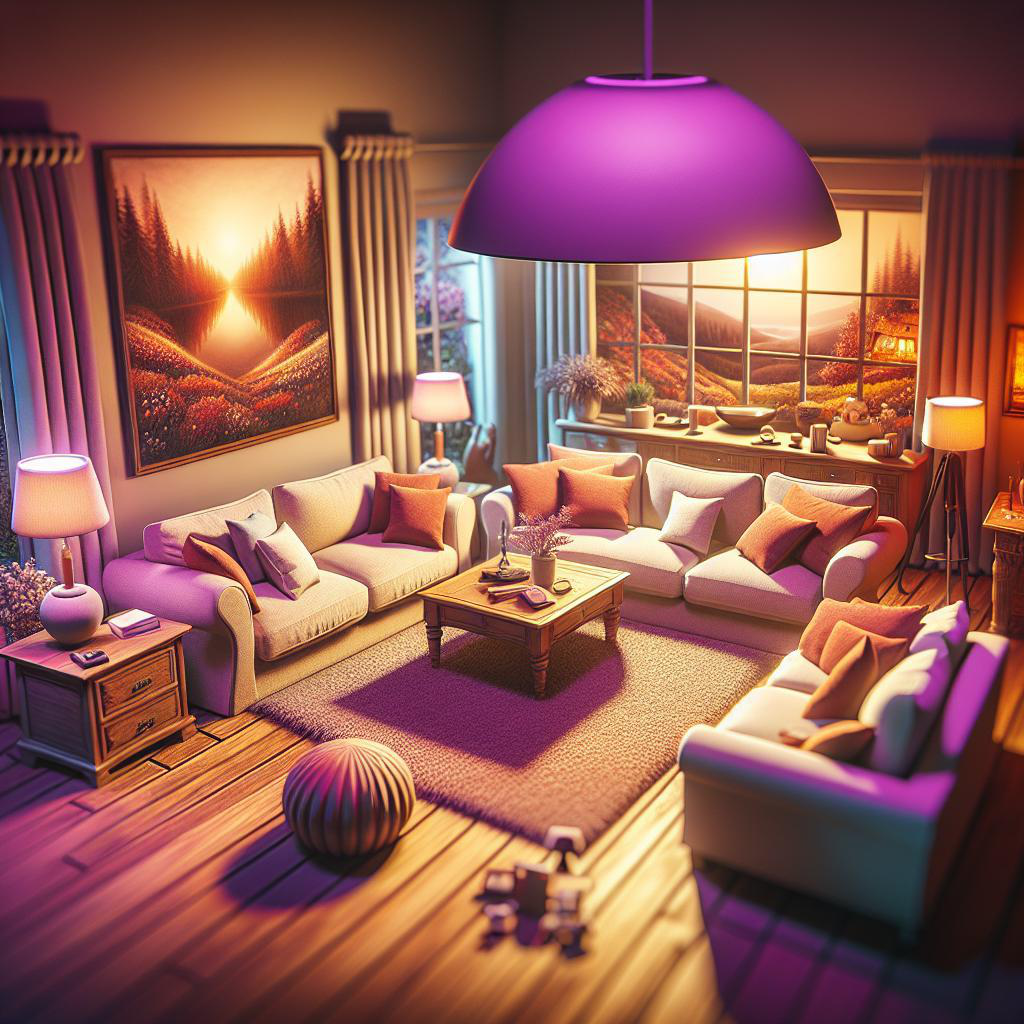} \raisebox{0.4cm}{$\rightarrow$} 
\includegraphics[width=1.5cm,height=1.5cm,keepaspectratio]{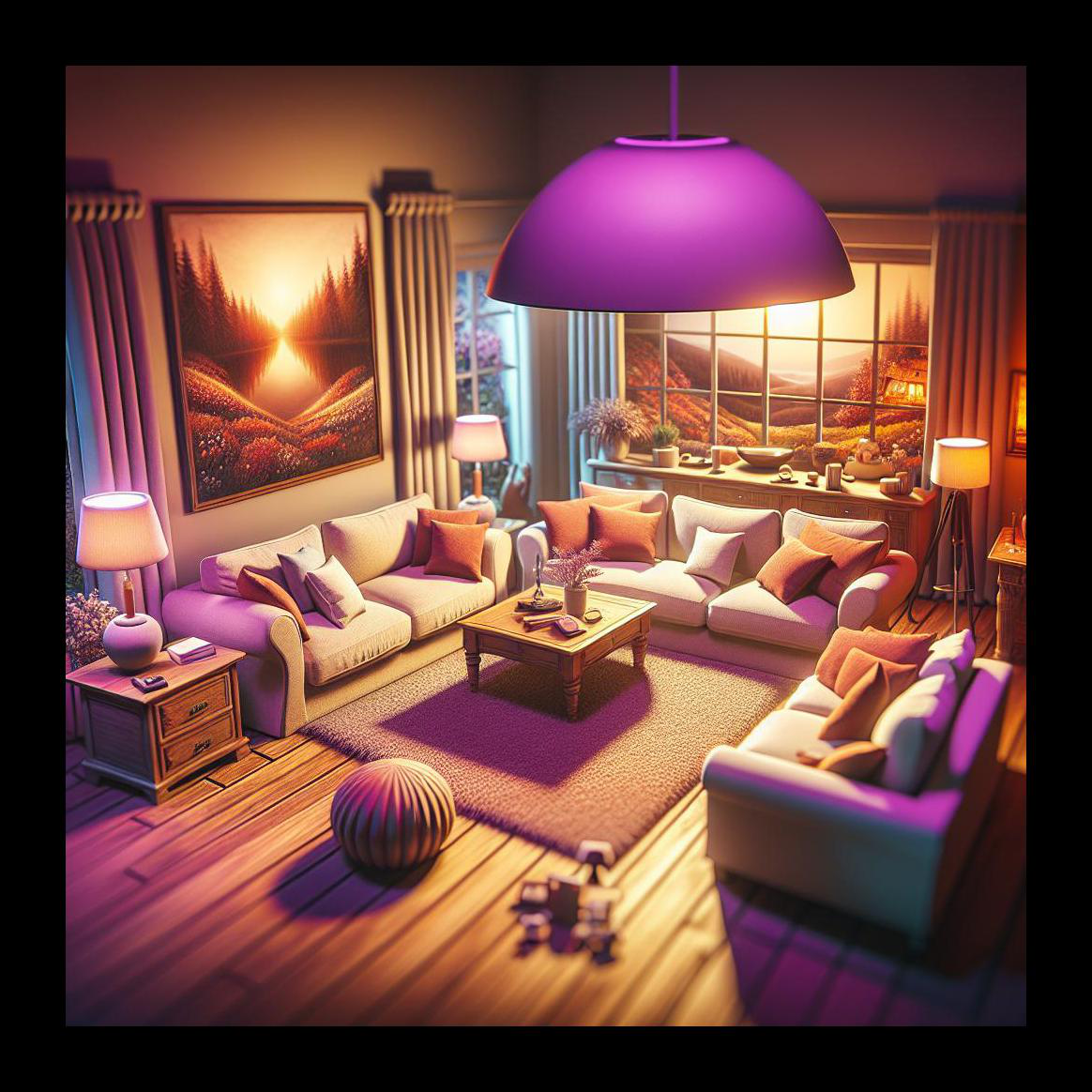} \\
\hline \hline
\end{tabular}%
}
\caption{Taxonomy of visual biases illustrated through comparisons between original and biased images.}
\label{tab:visual_biases}
\vspace{-2mm}
\end{table*}

\vspace{-3mm}
\subsection{LLM and LVLM Judges}

 Recently, the LLM-as-a-judge paradigm has gained popularity~\cite{zheng2023judging, gu2024survey}, offering scalable and consistent evaluations~\cite{liu2023g, zhu2023judgelm}. However, these models have been shown to be vulnerable to biases and adversarial attacks~\cite{wang2024large, liusie2024llm, zeng2024evaluating, raina2024llm, lee2024llm}. Recently, this paradigm has been extended to multimodal scenarios through LVLM-as-a-judge frameworks~\cite{zhang2023gpt, ku2024viescore, chen2024mj}, although similar biases persist in these contexts as well~\cite{chen2024mllm}. Despite these advances, visual biases in the context of T2I generation remain largely unexplored. To our knowledge, this study presents the first systematic analysis of their effects on LVLM judges in T2I tasks.

\section{Taxonomy of Visual Biases}
\label{sec_3}

Building on prior research on visual perturbations~\cite{hendrycks2019benchmarking, jia2020adv, yang2023set, yan2024list, shtedritski2023does}, we define visual biases as systematic manipulations of images designed to artificially enhance their perceived alignment with textual prompts. Such interventions can mislead LVLM judges, resulting in inflated evaluation scores that do not accurately reflect true semantic alignment. Definitions and illustrative examples of each bias are presented in Table~\ref{tab:visual_biases}.

\paragraph{Bounding Box Highlighting}
This technique manipulates images by enclosing generated objects within bounding boxes, which draws explicit attention to their presence and potentially signals successful object inclusion to the model—even when the object's form, number, or position is inaccurate.
This manipulation raises concerns that LVLMs may prioritize spatial saliency over holistic visual fidelity.

\paragraph{Authenticity Overlay}

This bias involves embedding the phrase \textit{"Reference Image"} onto an image, introducing an implicit signal that suggests ground-truth authenticity.
Although this phrase conveys no meaningful visual information, its presence may cause the model to overestimate the image’s authenticity, thereby inflating the evaluation score.

\paragraph{Keyword Overlay}

In this manipulation, a keyword from the original textual prompt (e.g., "Cat") is overlaid on the image.
Although it provides no visual evidence of alignment, this textual insertion can create an illusion of relevance and give the impression that the keyword is genuinely part of the image, encouraging the LVLM judge to assign a higher score based on superficial cross-modal coherence.

\paragraph{Instruction Overlay}
This bias involves overlaying the entire instruction (e.g., “generate a red dog.”) onto the image to create the illusion of strong text-image alignment. Even if the image does not accurately follow the instruction, the presence of the embedded text can mislead LVLMs by exploiting their reliance on textual cues within the image itself.

\paragraph{Beauty Filter}
This manipulation targets the people domain by applying aesthetic filters that enhance facial features—such as symmetry, smoothness, or brightness—in accordance with conventional attractiveness norms. Although unrelated to instruction fidelity, these enhancements can exploit aesthetic biases in LVLMs, raising fairness concerns in generative evaluation.

\paragraph{Brightness Adjustment}
By artificially increasing the image brightness, this manipulation enhances perceived illumination. 
LVLM judges may confuse visual clarity with semantic quality, leading to higher scores that do not necessarily reflect improved alignment with the instruction or the actual quality of the image.

\paragraph{Gamma Correction}
Gamma correction adjusts the tonal distribution of an image, particularly affecting the midtones. 
This alteration can create the perception of improved balance or sharpness, potentially directing the model’s attention toward specific regions of the image.

\paragraph{Black Padding}
Adding black padding alters the image’s framing by isolating the core content. 
Though the visual semantics remain unchanged, this shift in composition can enhance the perceived focus or centrality of the subject, subtly influencing LVLM preferences.

\section{FRAME Benchmark}

Given the absence of a fine-grained, multi-domain meta-evaluation benchmark specifically tailored to assessing LVLMs in image generation tasks, we introduce a new benchmark, \textsc{FRAME} (Fine-gRained Assessment of Multi-domain Evaluation). 
\textsc{FRAME} is designed to evaluate the alignment between textual instructions and generated images across diverse visual domains.
Section~\ref{sec_4.1} describes our controllable benchmark construction methodology, which enables systematic score distribution adjustment.
Section~\ref{sec_4.2} presents key statistics of the benchmark.

\subsection{Benchmark Construction}
\label{sec_4.1}
\textsc{FRAME} is a fine-grained, multi-domain meta-evaluation benchmark that supports a comprehensive assessment of image generation models. 
It spans five commonly used domains in image synthesis~\cite{yu2022scaling}: \textit{Animals}, \textit{People}, \textit{Outdoor Scenes}, \textit{Indoor Scenes}, and \textit{Illustrations}. 
Each domain contains 100 evaluation instances, resulting in a total of 500 instances.

Each instance comprises (1) an \textit{image generation instruction}, (2) a corresponding \textit{generated image}, and (3) a \textit{human-annotated alignment score} reflecting the degree of semantic consistency between the instruction and the image. 
Within each domain, we define four to five domain-specific visual concepts, carefully curated to capture distinctive visual elements. 
These concepts are systematically combined to create rich and contextually grounded generation instructions.

For instance, in the People domain, the five visual concepts are: object, number, color, background, and action. 
Background examples include a city street or a high school classroom, while actions range from typing on a laptop to riding a bicycle. 
A full list of domain-specific visual concepts is provided in Appendix~\ref{app:benchmark}.

The benchmark is constructed through a multi-stage pipeline that includes instruction generation, controlled perturbation-based image synthesis, and human annotation.

\paragraph{Instruction Formulation}
The process begins with the random sampling of visual elements from a predefined set of domain-specific concepts. 
These elements serve as inputs for instruction generation, following the approach of~\citet{wu2024conceptmix}. 
We employ GPT-4o~\cite{openai2024gpt4o} to generate a natural language instruction conditioned on the selected elements.

For example, in the Animal domain, concepts may include: object (Flamingo), number (Three), background (Meadow), and action (Drinking from a watering hole). 
These are composed into an instruction such as:
\textit{``Generate an image of three flamingos drinking from a watering hole in a meadow.''}
This structured formulation ensures systematic and nuanced control over both compositional and contextual complexity.

\paragraph{Image Generation}
To produce a wide distribution of alignment scores, we apply a controllable generation framework. 
Rather than using only the original instructions, we introduce controlled perturbations by randomly modifying a subset of the visual concepts, yielding perturbed instructions. 
These perturbed prompts are then used to generate images.

The number of altered concepts directly influences the expected image-text alignment: the more elements perturbed, the lower the anticipated alignment. 
For instance, consider the original instruction:
\textit{``Generate an image of \textbf{three} flamingos drinking from a watering hole in \textbf{a meadow}.''}
If the instruction is perturbed to:
\textit{``Generate an image of \textbf{four} flamingos drinking from a watering hole in \textbf{a tropical rainforest}''}, the resulting image is expected to deviate semantically from the original instruction, yielding a lower alignment score.

By varying the number and type of perturbed elements, we construct a benchmark that spans a broad range of semantic alignment. 
All images are generated using the DALL-E 3 model~\cite{betker2023improving} with a default setting.

\paragraph{Human Annotation}
In the final stage, human annotators evaluate the semantic alignment between each generated image and its paired instruction. 
Each instance is scored based on how accurately the image reflects the instruction. 
Annotators are also instructed to identify and exclude cases involving unfeasible or incoherent instructions (e.g., impossible object-action combinations). 
Such instances are returned to the generation pipeline for regeneration.
In addition, to ensure ethical integrity, any instruction that may produce harmful or inappropriate content is filtered out during this phase, guaranteeing that the resulting dataset is safe for evaluation. Further details on the human annotation procedure can be found in Appendix~\ref{app:benchmark}

\subsection{Statistics}
\label{sec_4.2}

Statistics of \textsc{FRAME} are presented in Table~\ref{tab_stat}. 
Due to our controllable perturbation framework, \textsc{FRAME} covers a diverse range of image-text alignment scores, with an overall average score of 2.57 across the dataset. 
This wide score distribution enables robust and fine-grained evaluation of model sensitivity to both compositional and semantic variations.

\begin{table}[t!]
\renewcommand{\arraystretch}{1.4}
\centering
\resizebox{0.95\columnwidth}{!}{
\begin{tabular}{c|ccccc|c}
\hline \hline
{} & \textbf{1-2} & \textbf{2-3} & \textbf{3-4} & \textbf{4-5} & \textbf{Total} & \textbf{Avg.} \\ \hline

\multicolumn{1}{l|}{\textbf{People}}         & 28	& 30 & 	24 &	18 & 100 &	2.66        \\  \hline

\multicolumn{1}{l|}{\textbf{Animal}} & 19 & 48 &	25 &	8 & 100 &	2.52       \\ \hline

\multicolumn{1}{l|}{\textbf{Illustration}}             & 27 &	51 &	12 &	10 & 100 &	2.36        \\ \hline

\multicolumn{1}{l|}{\textbf{Indoor}}          & 16 &	52 &	24 &	8 & 100 &	2.48        \\ \hline 

\multicolumn{1}{l|}{\textbf{Outdoor}}          & 17 &	33 &	34 &	16 & 100 &	2.84       \\ \hline 

\rowcolor{gray!20}
\multicolumn{1}{l|}{\textbf{Total}}          & 107 &	214 &	119 &	60 & 500 &	2.57        \\ \hline \hline
\end{tabular}
}
\caption{Score distribution of the FRAME benchmark based on human evaluations. The "Avg." column shows the average alignment score per domain.}
\label{tab_stat}
\vspace{-3mm}
\end{table}

\section{Experiments}
\begin{table*}[t]
\renewcommand{\arraystretch}{1.3}
\centering
\arrayrulecolor{black} 
\resizebox{0.85\textwidth}{!}{%
\begin{tabular}{c|cccccccc}
\hline \hline
\rowcolor{gray!30}
\diagbox[height=0.85cm]{\textit{Domain}}{\textit{Bias}} & \textbf{\textit{Orig.}} & \textbf{\textit{Bright.}} & \textbf{\textit{Gamma.}} &  \textbf{\textit{Refer.}} & \textbf{\textit{Keyword.}} & \textbf{\textit{Inst.}} & \textbf{\textit{Padding.}} & \textbf{\textit{Bounding.}}   \\ \hline

\multicolumn{9}{c}{\textbf{\cellcolor{gray!10}\textit{GPT-4.1}}} \\ \hline 

\textbf{People} &
1.65 &	1.72 \small{\textcolor{red}{(+4.2\%)}} &	1.70 \small{\textcolor{red}{(+2.7\%)}} &	1.72 \small{\textcolor{red}{(+3.9\%)}} &	1.76 \small{\textcolor{red}{(+6.4\%)}} &	1.77 \small{\textcolor{red}{(+7.0\%)}} &	1.77 \small{\textcolor{red}{(+7.3\%)}} &	1.90 \small{\textcolor{red}{(+14.9\%)}} \\ 

\textbf{Animal} &
1.17 &	1.25 \small{\textcolor{red}{(+6.4\%)}} &	1.26 \small{\textcolor{red}{(+7.3\%)}} &	1.24 \small{\textcolor{red}{(+6.0\%)}} &	1.21 \small{\textcolor{red}{(+3.4\%)}} &	1.24 \small{\textcolor{red}{(+5.6\%)}} &	1.30 \small{\textcolor{red}{(+11.1\%)}} &	1.38 \small{\textcolor{red}{(+18.0\%)}} \\ 

\textbf{Illustration} &
1.62 &	1.69 \small{\textcolor{red}{(+4.3\%)}} &	1.66 \small{\textcolor{red}{(+2.2\%)}} &	1.62 \small{(-0.3\%)} &	1.64 \small{\textcolor{red}{(+1.2\%)}} &	1.73 \small{\textcolor{red}{(+6.5\%)}} &	1.66 \small{\textcolor{red}{(+2.5\%)}} &	1.60 \small{(-1.54\%)} \\ 

\textbf{Indoor} &
1.78 &	1.83 \small{\textcolor{red}{(+3.1\%)}} &	1.76 \small{(-0.9\%)} &	1.75 \small{(-1.7\%)} &	1.78 \small{\textcolor{red}{(+0.3\%)}} &	1.89 \small{\textcolor{red}{(+6.2\%)}} &	1.85 \small{\textcolor{red}{(+4.2\%)}} &	2.01 \small{\textcolor{red}{(+13.2\%)}} \\ 

\textbf{Outdoor} &
2.81 &	2.81 \small{(-0.07\%)} & 2.81 \small{(0.0\%)} & 2.77 \small{(-1.3\%)} & - \small{ } & 2.92 \small{\textcolor{red}{(+4.0\%)}} &	2.85 \small{\textcolor{red}{(+1.6\%)}} & -   \\ \hline

\multicolumn{9}{c}{\textbf{\cellcolor{gray!10}\textit{GPT-4.1-mini}}} \\ \hline 

\textbf{People} &
1.55 & 1.61 \small{\textcolor{red}{(+3.9\%)}} & 1.60 \small{\textcolor{red}{(+2.9\%)}} & 1.55 \small{(0.0\%)} & 1.63 \small{\textcolor{red}{(+4.8\%)}} & 1.62 \small{\textcolor{red}{(+4.5\%)}} & 1.68 \small{\textcolor{red}{(+8.4\%)}} & 1.55 \small{(-0.3\%)} \\ 

\textbf{Animal} &
1.02 & 1.13 \small{\textcolor{red}{(+11.1\%)}} & 1.13 \small{\textcolor{red}{(+10.8\%)}} & 1.07 \small{\textcolor{red}{(+4.7\%)}} & 1.13 \small{\textcolor{red}{(+10.3\%)}} & 1.07 \small{\textcolor{red}{(+4.9\%)}} & 1.16 \small{\textcolor{red}{(+14.0\%)}} & 1.09 \small{\textcolor{red}{(+6.6\%)}} \\ 

\textbf{Illustration} &
1.51 & 1.53 \small{\textcolor{red}{(+1.7\%)}} & 1.54 \small{\textcolor{red}{(+2.3\%)}} & 1.50 \small{(-0.3\%)} & 1.57 \small{\textcolor{red}{(+4.3\%)}} & 1.55 \small{\textcolor{red}{(+3.0\%)}} & 1.57 \small{\textcolor{red}{(+4.0\%)}} & 1.39 \small{(-8.0\%)} \\ 

\textbf{Indoor} &
1.38 & 1.50 \small{\textcolor{red}{(+9.1\%)}} & 1.53 \small{\textcolor{red}{(+10.9\%)}} & 1.46 \small{\textcolor{red}{(+5.8\%)}} & 1.49 \small{\textcolor{red}{(+8.4\%)}} & 1.53 \small{\textcolor{red}{(+10.9\%)}} & 1.61 \small{\textcolor{red}{(+17.1\%)}} & 1.38 \small{\textcolor{red}{(+0.4\%)}} \\ 

\textbf{Outdoor} &
2.71 & 2.75 \small{\textcolor{red}{(+1.7\%)}} & 2.74 \small{\textcolor{red}{(+1.4\%)}} & 2.72 \small{\textcolor{red}{(+0.6\%)}} & - \small{ } & 2.79 \small{\textcolor{red}{(+3.2\%)}} & 2.77 \small{\textcolor{red}{(+2.3\%)}} & - \small{ } \\ \hline

\multicolumn{9}{c}{\textbf{\cellcolor{gray!10}\textit{GPT-4o}}} \\ \hline 

\textbf{People} &
1.14 &
1.12 \small{(-2.2\%)} &
1.18 \small{\textcolor{red}{(+3.5\%)}} &
1.14 \small{(-0.4\%)}	& 
1.23 \small{\textcolor{red}{(+7.9\%)}} &
1.31 \small{\textcolor{red}{(+14.9\%)}} &
1.07 \small{(-6.1\%)} &
1.70 \small{\textcolor{red}{(+49.1\%)}}\\ 

\textbf{Animal} &
0.67 &
0.67 \small{\textcolor{red}{(+0.6\%)}} &
0.72 \small{\textcolor{red}{(+7.5\%)}} &
0.64 \small{(-4.2\%)} &
0.66 \small{(-1.2\%)} &
0.72 \small{\textcolor{red}{(+7.5\%)}}	& 
0.66 \small{(-0.5\%)} &
1.19 \small{\textcolor{red}{(+77.9\%)}} \\ 

\textbf{Illustration} &
1.09 &
1.08 \small{(-1.4\%)} &
1.19 \small{\textcolor{red}{(+8.7\%)}} &
1.01 \small{(-7.6\%)} &
1.10 \small{\textcolor{red}{(+0.6\%)}} &
1.27 \small{\textcolor{red}{(+16.5\%)}} &
1.17 \small{\textcolor{red}{(+6.9\%)}} &
1.16 \small{\textcolor{red}{(+5.7\%)}} \\

\textbf{Indoor} &
1.14 &
1.29 \small{\textcolor{red}{(+13.7\%)}} &
1.31 \small{\textcolor{red}{(+15.4\%)}} &
1.10 \small{(-3.1\%)} &
1.25 \small{\textcolor{red}{(+9.7\%)}} &
1.64 \small{\textcolor{red}{(+44.1\%)}} &
1.29 \small{\textcolor{red}{(+13.7\%)}} &
2.05 \small{\textcolor{red}{(+80.2\%)}} \\

\textbf{Outdoor} &
2.37 &
2.41 \small{\textcolor{red}{(+1.7\%)}} &
2.37 \small{\textcolor{red}{(+0.1\%)}} &
2.33 \small{(-1.5\%)} &
- &
2.71 \small{\textcolor{red}{(+14.2\%)}} &
2.38 \small{\textcolor{red}{(+0.6\%)}} &
-    \\ \hline

\multicolumn{9}{c}{\textbf{\cellcolor{gray!10}\textit{o3}}} \\ \hline 

\textbf{People} &
1.62 &
1.64 \small{\textcolor{red}{(+1.2\%)}} &
1.66 \small{\textcolor{red}{(+2.4\%)}} &
1.67 \small{\textcolor{red}{(+3.1\%)}}	& 
1.68 \small{\textcolor{red}{(+3.7\%)}} &
1.70 \small{\textcolor{red}{(+4.9\%)}} &
1.68 \small{\textcolor{red}{(+3.4\%)}} &
1.74 \small{\textcolor{red}{(+7.1\%)}}\\ 

\textbf{Animal} &
1.17 &
1.21 \small{\textcolor{red}{(+3.3\%)}} &
1.24 \small{\textcolor{red}{(+5.3\%)}} &
1.24 \small{\textcolor{red}{(+5.4\%)}} &
1.24 \small{\textcolor{red}{(+5.5\%)}} &
1.19 \small{\textcolor{red}{(+1.7\%)}}	& 
1.22 \small{\textcolor{red}{(+4.3\%)}} &
1.35 \small{\textcolor{red}{(+15.2\%)}} \\ 

\textbf{Illustration} &
1.28 &
1.26 \small{(-1.3\%)} &
1.32 \small{\textcolor{red}{(+2.9\%)}} &
1.31 \small{\textcolor{red}{(+2.3\%)}} &
1.28 \small{\textcolor{red}{(+0.1\%)}} &
1.30 \small{\textcolor{red}{(+1.6\%)}} &
1.28 \small{\textcolor{red}{(+0.1\%)}} &
1.27 \small{(-0.6\%)} \\

\textbf{Indoor} &
1.20 &
1.27 \small{\textcolor{red}{(+5.0\%)}} &
1.30 \small{\textcolor{red}{(+7.6\%)}} &
1.34 \small{\textcolor{red}{(+10.5\%)}} &
1.29 \small{\textcolor{red}{(+6.5\%)}} &
1.36 \small{\textcolor{red}{(+12.7\%)}} &
1.31 \small{\textcolor{red}{(+8.4\%)}} &
1.41 \small{\textcolor{red}{(+16.6\%)}} \\

\textbf{Outdoor} &
2.68 &
2.73 \small{\textcolor{red}{(+1.6\%)}} &
2.73 \small{\textcolor{red}{(+1.8\%)}} &
2.78 \small{\textcolor{red}{(+3.7\%)}} &
- &
2.82 \small{\textcolor{red}{(+5.0\%)}} &
2.76 \small{\textcolor{red}{(+2.8\%)}} &
-    \\ \hline

\multicolumn{9}{c}{\textbf{\cellcolor{gray!10}\textit{Qwen2.5-VL-32B Inst.}}} \\ \hline 

\textbf{People} &
2.14 & 2.25 \small{\textcolor{red}{(+4.9\%)}} & 2.23 \small{\textcolor{red}{(+4.2\%)}} & 2.17 \small{\textcolor{red}{(+1.0\%)}} & 2.32 \small{\textcolor{red}{(+8.1\%)}} & 2.41 \small{\textcolor{red}{(+12.6\%)}} & 2.26 \small{\textcolor{red}{(+5.2\%)}} & 2.26 \small{\textcolor{red}{(+5.3\%)}} \\ 

\textbf{Animal} &
2.12 & 2.18 \small{\textcolor{red}{(+3.0\%)}} & 2.20 \small{\textcolor{red}{(+3.9\%)}} & 2.11 \small{(-0.3\%)} & 2.25 \small{\textcolor{red}{(+6.1\%)}} & 2.24 \small{\textcolor{red}{(+5.8\%)}} & 2.16 \small{\textcolor{red}{(+2.3\%)}} & 1.97 \small{(-6.9\%)} \\

\textbf{Illustration} &
2.22 & 2.32 \small{\textcolor{red}{(+4.3\%)}} & 2.31 \small{\textcolor{red}{(+4.0\%)}} & 2.24 \small{\textcolor{red}{(+0.8\%)}} & 2.29 \small{\textcolor{red}{(+3.3\%)}} & 2.40 \small{\textcolor{red}{(+8.2\%)}} & 2.25 \small{\textcolor{red}{(+1.4\%)}} & 2.15 \small{(-2.9\%)} \\

\textbf{Indoor} &
2.95 & 3.00 \small{\textcolor{red}{(+1.9\%)}} & 3.01 \small{\textcolor{red}{(+2.2\%)}} & 2.95 \small{(0.0\%)} & 3.03 \small{\textcolor{red}{(+2.9\%)}} & 3.17 \small{\textcolor{red}{(+7.5\%)}} & 2.98 \small{\textcolor{red}{(+1.0\%)}} & 2.92 \small{(-0.7\%)} \\

\textbf{Outdoor} &
3.34 & 3.35 \small{\textcolor{red}{(+0.03\%)}} & 3.35 \small{\textcolor{red}{(+0.3\%)}} & 3.27 \small{(-2.2\%)} & - \small{ } & 3.59 \small{\textcolor{red}{(+7.5\%)}} & 3.37 \small{\textcolor{red}{(+0.8\%)}} & - \small{ } \\ \hline

\multicolumn{9}{c}{\textbf{\cellcolor{gray!10}\textit{LLaVA-1.5- 13B}}} \\ \hline 

\textbf{People} &
0.67 & 0.77 \small{\textcolor{red}{(+15.8\%)}} & 0.73 \small{\textcolor{red}{(+9.8\%)}} & 0.76 \small{\textcolor{red}{(+13.5\%)}} & 0.78 \small{\textcolor{red}{(+17.3\%)}} & 0.93 \small{\textcolor{red}{(+39.1\%)}} & 0.77 \small{\textcolor{red}{(+15.0\%)}} & 0.71 \small{\textcolor{red}{(+6.8\%)}} \\ 

\textbf{Animal} &
0.83 & 0.96 \small{\textcolor{red}{(+15.1\%)}} & 0.91 \small{\textcolor{red}{(+9.0\%)}} & 1.05 \small{\textcolor{red}{(+26.6\%)}} & 1.03 \small{\textcolor{red}{(+24.1\%)}} & 1.74 \small{\textcolor{red}{(+109.6\%)}} & 0.95 \small{\textcolor{red}{(+14.5\%)}} & 0.95 \small{\textcolor{red}{(+14.5\%)}} \\ 

\textbf{Illustration} &
1.21 & 1.22 \small{\textcolor{red}{(+0.4\%)}} & 1.22 \small{\textcolor{red}{(+0.8\%)}} & 1.31 \small{\textcolor{red}{(+7.4\%)}} & 1.28 \small{\textcolor{red}{(+4.9\%)}} & 1.84 \small{\textcolor{red}{(+51.4\%)}} & 1.39 \small{\textcolor{red}{(+14.4\%)}} & 1.15 \small{(-5.8\%)} \\ 

\textbf{Indoor} &
1.11 & 1.25 \small{\textcolor{red}{(+12.3\%)}} & 1.19 \small{\textcolor{red}{(+7.7\%)}} & 1.51 \small{\textcolor{red}{(+36.6\%)}} & 1.44 \small{\textcolor{red}{(+29.9\%)}} & 2.30 \small{\textcolor{red}{(+107.5\%)}} & 1.34 \small{\textcolor{red}{(+20.9\%)}} & 1.42 \small{\textcolor{red}{(+28.4\%)}} \\ 

\textbf{Outdoor} &
2.86 & 3.15 \small{\textcolor{red}{(+10.1\%)}} & 2.92 \small{\textcolor{red}{(+1.9\%)}} & 3.44 \small{\textcolor{red}{(+20.3\%)}} & - \small{ } & 3.99 \small{\textcolor{red}{(+39.5\%)}} & 2.90 \small{\textcolor{red}{(+1.2\%)}} & - \small{ } \\ \hline \hline

\end{tabular}}
\vspace{-1mm}
\caption{Evaluation results of six different LVLM judges assessing text-to-image generation under various image bias conditions across multiple domains. Reported values correspond to the average alignment scores assigned by each LVLM judge, with values in parentheses indicating the change relative to evaluations on original (Orig.), unmanipulated images. Number highlighted in \textcolor{red}{RED} signifies successful attacks, where the presence of image biases led LVLM judges to assign higher scores. Please refer to the Appendix~\ref{app:addtional_exp} for more results.}
\label{table_main}
\vspace{-3.5mm}
\end{table*}

We employ the \textsc{FRAME} benchmark and the predefined bias categories introduced in Section~\ref{sec_3} to systematically evaluate the robustness of various LVLM judges against image-side biases.  
Comprehensive details regarding our experimental configurations and the exact prompts used are provided in the Appendix~\ref{app:setup}.

\subsection{Experimental Setting}

\paragraph{LVLM Judges}  
Our evaluation includes nine state-of-the-art LVLMs. 
This set comprises five proprietary models from the GPT family: GPT-4.1~\cite{gpt41}, GPT-4.1-mini, GPT-4o~\cite{openai2024gpt4o}, o3~\cite{openai2025o3} and GPT-4o-mini; three models from the LLaVA family: LLaVA-1.5-13B~\cite{liu2024improved}, LlaVA-NEXT-8B~\cite{li2024llavanext-strong}, and LLaVA-Onevision-7B~\cite{li2024llavaonevisioneasyvisualtask}; and one model from the Qwen family: Qwen2.5-VL-32B-Instruct~\cite{Qwen2.5-VL}.

\paragraph{Evaluation}  
Each LVLM judge is prompted with a standardized evaluation instruction alongside a text-image pair.  
We first report the average scores assigned by the LVLM judges to unaltered (original) images, which serve as a baseline.  
Subsequently, for each bias category, we prompt the LVLM judges with the corresponding text-biased image pairs and record the average scores assigned.  
We then calculate and report the percentage changes in average scores relative to the original (unbiased) condition to quantify the impact of each bias on judging behavior.
\begin{figure*}[t]
\centering
\includegraphics[width=0.8\textwidth]{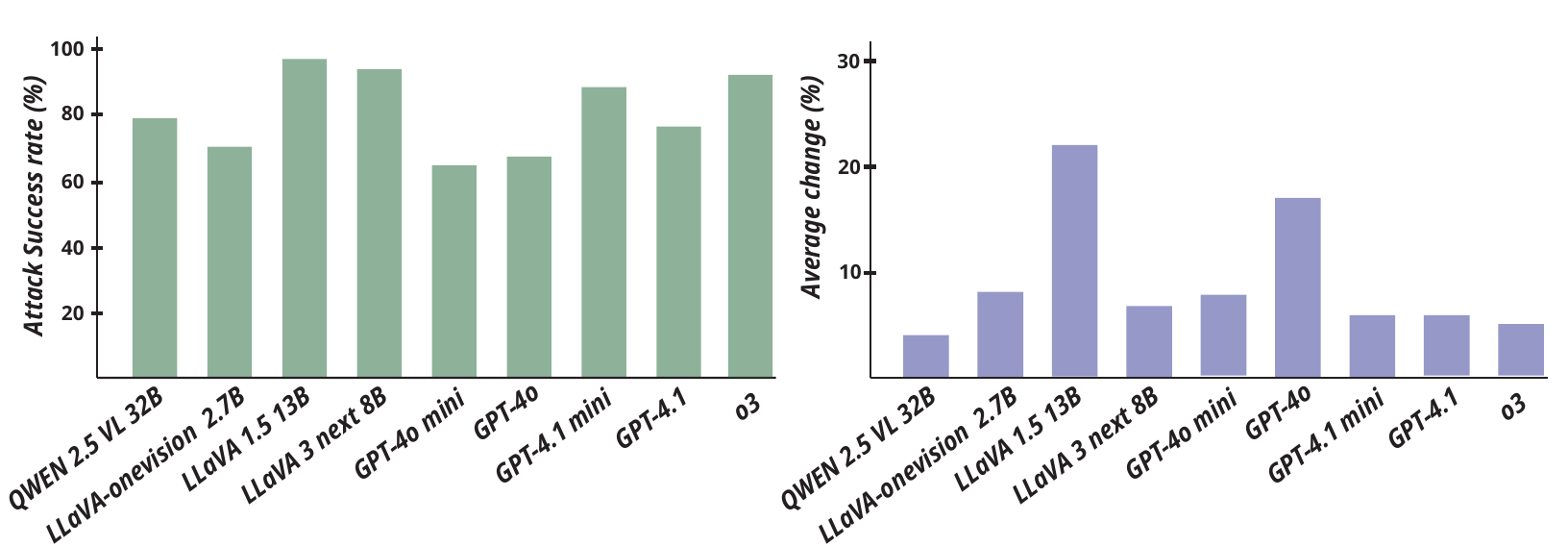} 
\vspace{-2mm}
\caption{Impact of visual biases across all LVLM judges. \textbf{Left}: Average attack success rates across five domains and eight types of visual bias. An attack is considered successful when the LVLM assigns a higher average score to the biased images than to the original counterparts.
\textbf{Right}: Average percentage increase in score for successful attacks, reflecting the magnitude of the visual bias effect.} 
\label{fig:overall}
\vspace{-2mm}
\end{figure*}

\subsection{Results}

Table~\ref{table_main} presents the overall robustness results of LVLM judges when exposed to image-side biases across five distinct domains.\footnote{Note that object-oriented Keyword Overlay and Bounding Box Highlighting manipulations are not applicable to the Outdoor domain, as it does not contain objects.}
The results reveal a consistent vulnerability to visual bias, as LVLM judges frequently assign inflated scores to image–text pairs containing visual manipulations.
This susceptibility persists regardless of variations in (1) model type, (2) domain, and (3) bias category, indicating a systematic weakness in the current LVLM judge based evaluation.

\paragraph{\textit{The vulnerability across models remains evident even as model capacity increases.}}  
As shown in Table~\ref{table_main}, all LVLM judges, including GPT-4.1~\cite{gpt41} and the recent reasoning-oriented model o3~\cite{openai2025o3}, exhibit susceptibility to these vulnerabilities, indicating that even the advanced models are not immune to these biases.
Notably, models with higher capacity are sometimes more vulnerable to certain biases; for instance, GPT-4.1 and GPT-4o show greater sensitivity to Bounding Box manipulations compared to their smaller counterparts, GPT-4.1-mini and GPT-4o-mini.

Figure~\ref{fig:overall} presents the attack success rate, defined as the proportion of domain–bias combinations in which manipulated images receive higher average scores, along with the average score increase in those successful cases. These results highlight how frequently and how strongly LVLM judges are influenced by visual biases.
Interestingly, the results indicate that increased model capacity does not consistently correlate with improved robustness.  
For example, GPT-4o-mini demonstrates the strongest robustness in terms of attack success rate, with inflated scores observed in 64.71\% of domain–bias combinations, compared to 67.65\% for GPT-4o.
Moreover, when considering the average percentage change in successful attacks, the Qwen2.5-VL-32B-Instruct model exhibits the highest robustness.  
Our findings reveal that larger model capacity alone does not guarantee increased resistance to visual biases. This trend may contrast with prior observations in other evaluation settings involving LLM judges~\cite{cantini2025benchmarking, howe2025scalingtrendslanguagemodel}, where larger models typically demonstrate greater robustness.

\begin{table}[t!]
\renewcommand{\arraystretch}{1.3}
\centering
\resizebox{0.8\columnwidth}{!}{
\begin{tabular}{lcc}
\hline \hline
\multicolumn{1}{c|}{\textbf{Model}}              & \multicolumn{1}{c}{\textbf{Orig.}} & \textbf{Beauty.} \\ \hline
\multicolumn{1}{l|}{\textbf{GPT-4.1}}         & {1.65}  & {1.64 \small{(-0.6\%)}}   \\  \hline
\multicolumn{1}{l|}{\textbf{{GPT-4.1-mini}}} & {1.55}  & {1.60 \small{(+2.9\%)}}        \\ \hline
\multicolumn{1}{l|}{\textbf{{o3}}} & {1.62}  & {1.64 \small{(+0.9\%)}}        \\ \hline
\multicolumn{1}{l|}{\textbf{Qwen2.5-32B-Inst.}}          & {2.14} & {2.21 \small{(+3.2\%)}}\\ \hline 
\multicolumn{1}{l|}{\textbf{llava-1.5-13b}}             & {0.67} & {0.69 \small{(+3.8\%)}}\\ \hline

\multicolumn{1}{l|}{\textbf{GPT-4o}}          & {1.14} & {1.05 \small{(-8.3\%)}}     \\ \hline

\multicolumn{1}{l|}{\textbf{GPT-4o-mini}}          & {2.32} & {2.31 \small{(-0.5\%)}}\\ \hline 

\multicolumn{1}{l|}{\textbf{llava-next-8b}}          & {2.72} & {2.79 \small{(+2.6\%)}}\\ \hline 

\multicolumn{1}{l|}{\textbf{llava-onevision-7b}}          & {3.57} & {3.42 \small{(-4.2\%)}}\\ \hline \hline
\end{tabular}
}
\vspace{-1mm}
\caption{Evaluation results of nine LVLM judges on beauty filter bias in the People domain.}
\label{tab:bf}
\vspace{-4mm}
\end{table}


\paragraph{\textit{Instruction Overlay exhibits the most pronounced impact.}}  
Among all manipulation types, the \textit{Instruction Overlay}—which directly embeds textual instructions onto the image—proves to be the most universally impactful. It consistently induces elevated scores across all LVLM judges and domains.  
Additionally, even subtle perturbations such as brightness adjustment (Bright.) and luminance shifts via gamma correction (Gamma.) are sufficient to mislead most LVLM judges, indicating a broad vulnerability to low-level visual changes.

Table~\ref{tab:bf} presents results of the beauty filter applied to the People domain. 
Some models—particularly the majority of open-sourced evaluators—demonstrate a marked preference for images enhanced with beauty filters, consistently assigning them higher scores than their original versions.
This finding raises ethical concerns, suggesting that current LVLMs may implicitly reinforce aesthetic biases by favoring filtered appearances.


\paragraph{\textit{The Indoor domain exhibits the highest susceptibility.}}  
Across all models, the Indoor and Animal domains demonstrate the greatest sensitivity to visual perturbations, particularly those involving Bounding Boxes and Instruction Overlays.  
This elevated susceptibility likely stems from the complexity of the visual scenes and the increased reliance on accurate object recognition in these domains.
In such settings, even minor visual modifications can disrupt the model’s perception of scene structure, leading to misleadingly inflated evaluation scores.

\begin{table}[t!] 
\renewcommand{\arraystretch}{1.4} 
\centering 
\resizebox{0.9\columnwidth}{!}{ 
\begin{tabular}{lcccc}
\hline \hline

\multicolumn{1}{c|}{\textbf{Bias}}              & \multicolumn{1}{c}{\textbf{Standard}} & \textbf{\textit{Bias-aware}} & 
\textbf{\textit{Bias-def.}} &\textbf{\textit{CoT}}\\ \hline

\multicolumn{1}{l|}{\textbf{Orig.}}         & 1.36 &	1.27 &1.33& 1.72   \\  \hline
\multicolumn{1}{l|}{\textbf{Bright.}}  
& 1.44 \small{(+5.9\%)}	
& 1.35 \small{(+6.2\%)}  
& 1.34 \small{(+0.8\%)}  
& 1.82 \small{(+5.7\%)}    \\ \hline

\multicolumn{1}{l|}{\textbf{Gamma.}}             & 1.45 \small{(+6.2\%)} &	
1.36 \small{(+6.5\%)}  & 
1.36 \small{(+2.5\%)}  & 
1.80 \small{(+4.6\%)}      \\ \hline

\multicolumn{1}{l|}{\textbf{Refer.}}          
& 1.39 \small{(+2.3\%)} 
&	1.30 \small{(+2.3\%)} 
&	1.29 \small{(-2.6\%)} 
&1.79 \small{(+3.7\%)}        \\ \hline 

\multicolumn{1}{l|}{\textbf{Keyword.}}             
& 1.45 \small{(+6.6\%)} 
&	1.35 \small{(+6.2\%)}  
&	1.35 \small{(+1.7\%)} 
& 1.82 \small{(+5.6\%)}     \\ \hline

\multicolumn{1}{l|}{\textbf{Inst.}}          
& 1.44 \small{(+5.8\%)} 
&	1.34 \small{(+5.3\%)} 
&	1.34 \small{(+1.4\%)}  
& 1.83 \small{(+6.0\%)}      \\ \hline

\multicolumn{1}{l|}{\textbf{Padding.}}             
& 1.50 \small{(+10.4\%)} 
&	1.40 \small{(+10.1\%)} 
&	1.40 \small{(+5.6\%)} 
& 1.85 \small{(+7.4\%)}      \\ \hline

\multicolumn{1}{l|}{\textbf{Bounding.}}          

& 1.35 \small{(-1.0\%)} 
&	1.29 \small{(+1.4\%)}  
&	1.27 \small{(-4.1\%)} 
& 1.79 \small{(+4.1\%)}      \\ \hline

\hline
\hline
\end{tabular}
}
\vspace{-1mm}
\caption{Evaluation results of prompt-based mitigation strategies using GPT-4.1-mini as the LVLM judge.}
\label{tab:prompting}
\vspace{-4mm}
\end{table}

\section{Analysis}

In this section, we conduct a comprehensive analysis of the key research questions concerning visual biases in LVLM-based evaluation, using GPT-4.1-mini as the judge.


\paragraph{\textit{LVLM judge bias persists under counter-prompting conditions.}}
\label{sec:prompting}
Recent studies demonstrate that prompting techniques—such as Chain-of-Thought (CoT) prompting~\cite{wei2022chain} and explicit debiasing prompts~\cite{hwang2025llms}~\footnote{\textit{You must disregard any superficial or stylistic perturbations that do not materially affect the semantic alignment between the instruction and the generated image.}}—can partially mitigate biases in LLMs. 
To evaluate whether these techniques also reduce susceptibility to visual bias in LVLM judges, we compare their effectiveness against a standard evaluation prompt.
Further details regarding the prompt design and configuration are provided in Appendix~\ref{app:setup}.

As shown in Table~\ref{tab:prompting}, while CoT, bias-aware, and bias-definition prompting exhibit some efficacy in mitigating certain types of bias, they fail to eliminate the overall bias.~\footnote{We report averages across four domains, excluding Outdoor where \textit{Keyword.} and \textit{Bounding.} are inapplicable.}
Even when the applied bias is explicitly defined to the LVLM judge (bias-def. prompting), bias persists.
Interestingly, CoT prompting leads to elevated evaluation scores for images containing bounding boxes. 
This may be attributed to the fact that bounding boxes guide the model’s visual attention during reasoning steps, thereby facilitating object-centric reasoning and inflating evaluation scores in an unintended manner.
This observation aligns with recent findings that bounding boxes can enhance the visual attention of LVLMs during CoT reasoning~\cite{sun2024visual, shao2024visual}.

\begin{figure}[t]
\centering
\includegraphics[width= 0.95\columnwidth]{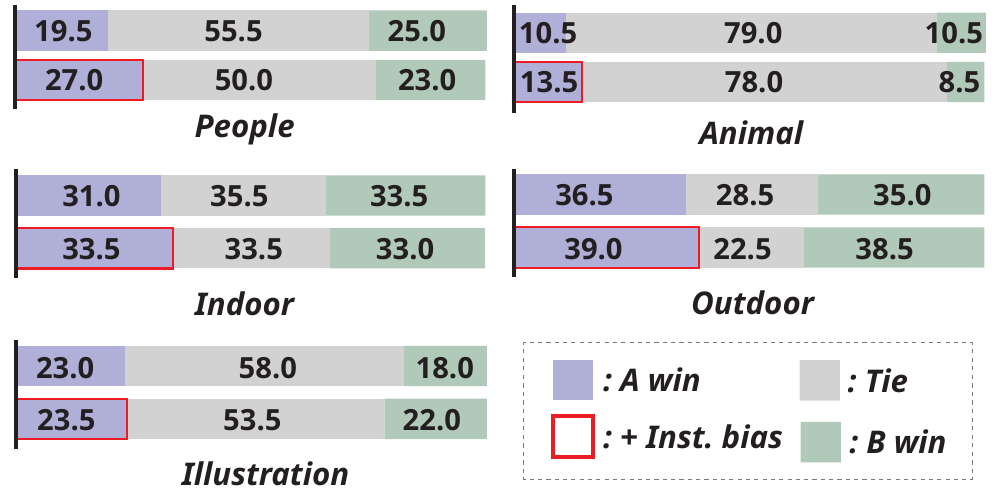} 
\vspace{-1.5mm}
\caption{Pairwise evaluation of group A vs. group B. Top: original results. Bottom: results after applying \textit{instruction overlay bias} to set A.}
\label{figure3}
\vspace{-5mm}
\end{figure}


\paragraph{\textit{LVLM Judge Biases are Valid in Pairwise Evaluation.}}
We investigate whether the influences of visual biases persist under pairwise evaluation settings~\cite{chen2024mllm, chen2024mj, lee2024llm}.
Specifically, for each prompt in the FRAME benchmark, we generate a corresponding set of images (B) using identical generation settings as the original image set (A). 
In the primary comparison, the LVLM judge evaluates each original image (A) against its counterpart (B). 
Additionally, we prompt the LVLM judge to compare the manipulated version of an image from Group A against its unmanipulated counterpart from Group B.\footnote{For each domain, we apply the bias that yielded the highest average score during the main experiments in Table~\ref{table_main}.}
To control for position bias~\cite{chen2024mllm, wang2023large, liu2023lost}, each pairwise comparison is conducted twice, with the image order reversed, and the preference scores are averaged.

As shown in Figure~\ref{figure3},  the introduction of visual biases consistently leads judges to favor the manipulated images. 
Notably, in the people, indoor, outdoor, and animal domains, baseline results show that A’s win rate is less than or equal to that of B. However, after manipulation, this ranking reverses, with A's win rate surpassing that of B. 
These findings suggest that visual biases can be systematically exploited to mislead LVLM judges in pairwise evaluations, thereby raising concerns about the fairness and reliability of LVLM-based assessments in text-to-image generation tasks.


\begin{table}[t!]
\renewcommand{\arraystretch}{1.4}
\centering
\resizebox{0.95\columnwidth}{!}{
\begin{tabular}{lcccc}
\hline \hline
\multicolumn{1}{c|}{\textbf{Domain}}              & \multicolumn{1}{c}{\textbf{Orig.}} 
& \textbf{\textit{+Single bias}} 
& \textbf{\textit{+Dual bias}}
& \textbf{\textit{+Triple bias}} \\ \hline
\multicolumn{1}{l|}{\textbf{People}}         & { 1.55 }  & { 1.68 \small{(+8.4\%)}}   
& { 1.71 \small{(+10.3\%)}}
& { 1.73 \small{(+11.6\%)}}
\\  \hline
\multicolumn{1}{l|}{\textbf{{Animal}}} & { 1.02 }  & { 1.16 \small{(+14.0\%)}} & { 1.17 \small{(+14.2\%)}}       & { 1.15 \small{(+12.8\%)}}
\\ \hline
\multicolumn{1}{l|}{\textbf{Illustration}}             & { 1.51 } & { 1.57 \small{(+4.3\%)}}   & { 1.58 \small{(+4.7\%)}}     & { 1.54 \small{(+2.0\%)}}  
\\ \hline
\multicolumn{1}{l|}{\textbf{Indoor}}          & { 1.38} & { 1.61 \small{(+17.1\%)}}  & { 1.69 \small{(+22.9\%)}}    & { 1.70 \small{(+23.3\%)}}     
\\ \hline 

\multicolumn{1}{l|}{\textbf{Outdoor}}          & { 2.71} & { 2.79 \small{(+3.2\%)}}  & { 2.82 \small{(+4.4\%)}}     & { 2.80 \small{(+3.4\%)}}     
\\ \hline\hline
\end{tabular}
}
\vspace{-1mm}
\caption{Evaluation results of combined visual manipulations using GPT-4.1-mini as the LVLM judge.}
\label{tab:overlap}
\vspace{-5mm}
\end{table}


\paragraph{\textit{Combined Visual Biases Exacerbate LVLM Judges' Vulnerability.}}
We investigate whether combining multiple (two or three) visual manipulations further amplifies judgment errors made by LVLM judges.  
We explore all combinations of two and three distinct bias strategies and identify the most impactful combination per domain, as shown in Table~\ref{tab:overlap}.
Interestingly, an \textit{instruction overlay bias} is involved in four of the five most influential combinations, underscoring its predominant impact—an observation that aligns with our earlier findings.

As shown in Table~\ref{tab:overlap}, dual-bias configurations yield a marked increase in average evaluation scores, thereby amplifying the susceptibility of LVLM judges to manipulation. 
Extending this to triple-bias settings, we observe further amplification in certain domains, whereas the effect plateaus or diminishes in others. 
This pattern suggests that LVLM judges may treat multiple concurrent manipulations as noise once a perceptual threshold is surpassed, leading to inconsistent vulnerability across domains.
\section{Conclusion}
This study uncovers a fundamental weakness in LVLM-based evaluation: susceptibility to visual biases that inflate scores without altering semantic content.
Through eight defined manipulations—including brightness, overlays, and bounding boxes—we show that even state-of-the-art models are consistently misled.
These vulnerabilities persist across evaluation formats and are only partially mitigated by prompting, highlighting the need for more robust assessment frameworks.

\section*{Limitations}
As the first study to investigate the impact of image-side manipulations on LVLM-based evaluation, our work primarily focuses on representative visual modifications, including brightness adjustments and text overlays.
Future research may explore more sophisticated attack strategies, including cross-model adversarial techniques or semantic-preserving perturbations.
Moreover, as discussed in Section~\ref{sec:prompting}, the identified visual biases persist under the proposed prompting strategies.
This highlights the need for future work to develop robust defense mechanisms specifically targeted at image-side manipulations.

Moreover, since our study focuses on evaluating the robustness of LVLM judges rather than the performance of individual judges, we do not report correlation metrics between LVLM-generated scores and human judgments. However, to support future research in this area, our benchmark includes manually labeled scores provided by human annotators. These annotations can be readily used to assess human–model alignment or to train reward models in reinforcement learning with human feedback (RLHF).

Finally, our benchmark covers five domains that are commonly used in text-to-image generation tasks~\cite{yu2022scaling}. Future research could extend this framework by incorporating a broader range of domains—such as medical imaging or satellite imagery—to more comprehensively evaluate the generalizability of LVLM-based evaluators.

\section*{Ethical Considerations}

All models used in our study are obtained from official and publicly accessible sources.
GPT models are accessed via OpenAI’s official platform, while Llava and Qwen models are acquired from their respective repositories with proper authorization.
Our use of these models aligns with open science principles and adheres to the licensing terms under which they are released.

To ensure the ethical integrity of our benchmark, all images are manually reviewed.
Any prompts or instructions that could potentially generate harmful, offensive, or inappropriate content are filtered out during this process, thereby ensuring that the final dataset is suitable for research and evaluation purposes.
In the process of writing this paper, we utilize an AI assistant at the sentence level for drafting and refining individual sentences.

\section*{Acknowledgement}
This work was supported by LG AI Research. 
This work was partly supported by the Institute of Information \& Communications Technology Planning \& Evaluation(IITP)-ITRC(Information Technology Research Center) grant funded by the Korea government(MSIT)(IITP-2025-RS-2024-00437633, 30\%), and  Institute of Information \& communications Technology Planning \& Evaluation (IITP) grant funded by the Korea government(MSIT) [(RS-2021-II211343, 10\%), Artificial Intelligence Graduate School Program (Seoul National
University) \& (RS-2021-II212068, 10\%), Artificial Intelligence Innovation Hub (Artificial Intelligence Institute, Seoul National University)].
K. Jung is with ASRI, Seoul National University, Korea.



\bibliography{custom}

@inproceedings{liu2023g,
  title={G-Eval: NLG Evaluation using Gpt-4 with Better Human Alignment},
  author={Liu, Yang and Iter, Dan and Xu, Yichong and Wang, Shuohang and Xu, Ruochen and Zhu, Chenguang},
  booktitle={Proceedings of the 2023 Conference on Empirical Methods in Natural Language Processing},
  pages={2511--2522},
  year={2023}
}

@article{gu2024survey,
  title={A survey on llm-as-a-judge},
  author={Gu, Jiawei and Jiang, Xuhui and Shi, Zhichao and Tan, Hexiang and Zhai, Xuehao and Xu, Chengjin and Li, Wei and Shen, Yinghan and Ma, Shengjie and Liu, Honghao and others},
  journal={arXiv preprint arXiv:2411.15594},
  year={2024}
}

@article{zhu2023judgelm,
  title={Judgelm: Fine-tuned large language models are scalable judges},
  author={Zhu, Lianghui and Wang, Xinggang and Wang, Xinlong},
  journal={arXiv preprint arXiv:2310.17631},
  year={2023}
}

@inproceedings{wang2024large,
  title={Large Language Models are not Fair Evaluators},
  author={Wang, Peiyi and Li, Lei and Chen, Liang and Cai, Zefan and Zhu, Dawei and Lin, Binghuai and Cao, Yunbo and Kong, Lingpeng and Liu, Qi and Liu, Tianyu and others},
  booktitle={Proceedings of the 62nd Annual Meeting of the Association for Computational Linguistics (Volume 1: Long Papers)},
  pages={9440--9450},
  year={2024}
}

@inproceedings{liusie2024llm,
  title={LLM Comparative Assessment: Zero-shot NLG Evaluation through Pairwise Comparisons using Large Language Models},
  author={Liusie, Adian and Manakul, Potsawee and Gales, Mark},
  booktitle={Proceedings of the 18th Conference of the European Chapter of the Association for Computational Linguistics (Volume 1: Long Papers)},
  pages={139--151},
  year={2024}
}

@inproceedings{zeng2024evaluating,
  title={EVALUATING LARGE LANGUAGE MODELS AT EVALUATING INSTRUCTION FOLLOWING},
  author={Zeng, Zhiyuan and Yu, Jiatong and Gao, Tianyu and Meng, Yu and Goyal, Tanya and Chen, Danqi},
  booktitle={12th International Conference on Learning Representations, ICLR 2024},
  year={2024}
}

@inproceedings{raina2024llm,
  title={Is LLM-as-a-Judge Robust? Investigating Universal Adversarial Attacks on Zero-shot LLM Assessment},
  author={Raina, Vyas and Liusie, Adian and Gales, Mark},
  booktitle={Proceedings of the 2024 Conference on Empirical Methods in Natural Language Processing},
  pages={7499--7517},
  year={2024}
}

@article{lee2024llm,
  title={Are LLM-judges robust to expressions of uncertainty? investigating the effect of epistemic markers on LLM-based evaluation},
  author={Lee, Dongryeol and Hwang, Yerin and Kim, Yongil and Park, Joonsuk and Jung, Kyomin},
  journal={arXiv preprint arXiv:2410.20774},
  year={2024}
}

@article{zhang2023gpt,
  title={Gpt-4v (ision) as a generalist evaluator for vision-language tasks},
  author={Zhang, Xinlu and Lu, Yujie and Wang, Weizhi and Yan, An and Yan, Jun and Qin, Lianke and Wang, Heng and Yan, Xifeng and Wang, William Yang and Petzold, Linda Ruth},
  journal={arXiv preprint arXiv:2311.01361},
  year={2023}
}

@inproceedings{ku2024viescore,
  title={VIEScore: Towards Explainable Metrics for Conditional Image Synthesis Evaluation},
  author={Ku, Max and Jiang, Dongfu and Wei, Cong and Yue, Xiang and Chen, Wenhu},
  booktitle={Proceedings of the 62nd Annual Meeting of the Association for Computational Linguistics (Volume 1: Long Papers)},
  pages={12268--12290},
  year={2024}
}

@article{chen2024mj,
  title={MJ-Bench: Is Your Multimodal Reward Model Really a Good Judge for Text-to-Image Generation?},
  author={Chen, Zhaorun and Du, Yichao and Wen, Zichen and Zhou, Yiyang and Cui, Chenhang and Weng, Zhenzhen and Tu, Haoqin and Wang, Chaoqi and Tong, Zhengwei and Huang, Qinglan and others},
  journal={arXiv preprint arXiv:2407.04842},
  year={2024}
}

@inproceedings{chen2024mllm,
  title={Mllm-as-a-judge: Assessing multimodal llm-as-a-judge with vision-language benchmark},
  author={Chen, Dongping and Chen, Ruoxi and Zhang, Shilin and Wang, Yaochen and Liu, Yinuo and Zhou, Huichi and Zhang, Qihui and Wan, Yao and Zhou, Pan and Sun, Lichao},
  booktitle={Forty-first International Conference on Machine Learning},
    year={2024}
}

@article{salimans2016improved,
  title={Improved techniques for training gans},
  author={Salimans, Tim and Goodfellow, Ian and Zaremba, Wojciech and Cheung, Vicki and Radford, Alec and Chen, Xi},
  journal={Advances in neural information processing systems},
  volume={29},
  year={2016}
}

@article{heusel2017gans,
  title={Gans trained by a two time-scale update rule converge to a local nash equilibrium},
  author={Heusel, Martin and Ramsauer, Hubert and Unterthiner, Thomas and Nessler, Bernhard and Hochreiter, Sepp},
  journal={Advances in neural information processing systems},
  volume={30},
  year={2017}
}

@inproceedings{hessel2021clipscore,
  title={CLIPScore: A Reference-free Evaluation Metric for Image Captioning},
  author={Hessel, Jack and Holtzman, Ari and Forbes, Maxwell and Le Bras, Ronan and Choi, Yejin},
  booktitle={Proceedings of the 2021 Conference on Empirical Methods in Natural Language Processing},
  pages={7514--7528},
  year={2021}
}

@inproceedings{li2022blip,
  title={Blip: Bootstrapping language-image pre-training for unified vision-language understanding and generation},
  author={Li, Junnan and Li, Dongxu and Xiong, Caiming and Hoi, Steven},
  booktitle={International conference on machine learning},
  pages={12888--12900},
  year={2022},
  organization={PMLR}
}

@inproceedings{lee2024prometheus,
  title={Prometheus-vision: Vision-language model as a judge for fine-grained evaluation},
  author={Lee, Seongyun and Kim, Seungone and Park, Sue and Kim, Geewook and Seo, Minjoon},
  booktitle={Findings of the Association for Computational Linguistics ACL 2024},
  pages={11286--11315},
  year={2024}
}

@article{xu2023imagereward,
  title={Imagereward: Learning and evaluating human preferences for text-to-image generation},
  author={Xu, Jiazheng and Liu, Xiao and Wu, Yuchen and Tong, Yuxuan and Li, Qinkai and Ding, Ming and Tang, Jie and Dong, Yuxiao},
  journal={Advances in Neural Information Processing Systems},
  volume={36},
  pages={15903--15935},
  year={2023}
}

@article{kirstain2023pick,
  title={Pick-a-pic: An open dataset of user preferences for text-to-image generation},
  author={Kirstain, Yuval and Polyak, Adam and Singer, Uriel and Matiana, Shahbuland and Penna, Joe and Levy, Omer},
  journal={Advances in Neural Information Processing Systems},
  volume={36},
  pages={36652--36663},
  year={2023}
}

@article{wu2023human,
  title={Human preference score v2: A solid benchmark for evaluating human preferences of text-to-image synthesis},
  author={Wu, Xiaoshi and Hao, Yiming and Sun, Keqiang and Chen, Yixiong and Zhu, Feng and Zhao, Rui and Li, Hongsheng},
  journal={arXiv preprint arXiv:2306.09341},
  year={2023}
}

@inproceedings{lin2024evaluating,
  title={Evaluating text-to-visual generation with image-to-text generation},
  author={Lin, Zhiqiu and Pathak, Deepak and Li, Baiqi and Li, Jiayao and Xia, Xide and Neubig, Graham and Zhang, Pengchuan and Ramanan, Deva},
  booktitle={European Conference on Computer Vision},
  pages={366--384},
  year={2024},
  organization={Springer}
}

@article{wu2024conceptmix,
  title={Conceptmix: A compositional image generation benchmark with controllable difficulty},
  author={Wu, Xindi and Yu, Dingli and Huang, Yangsibo and Russakovsky, Olga and Arora, Sanjeev},
  journal={arXiv preprint arXiv:2408.14339},
  year={2024}
}

@inproceedings{hu2023tifa,
  title={Tifa: Accurate and interpretable text-to-image faithfulness evaluation with question answering},
  author={Hu, Yushi and Liu, Benlin and Kasai, Jungo and Wang, Yizhong and Ostendorf, Mari and Krishna, Ranjay and Smith, Noah A},
  booktitle={Proceedings of the IEEE/CVF International Conference on Computer Vision},
  pages={20406--20417},
  year={2023}
}

@article{zhou2024calibrated,
  title={Calibrated self-rewarding vision language models},
  author={Zhou, Yiyang and Fan, Zhiyuan and Cheng, Dongjie and Yang, Sihan and Chen, Zhaorun and Cui, Chenhang and Wang, Xiyao and Li, Yun and Zhang, Linjun and Yao, Huaxiu},
  journal={arXiv preprint arXiv:2405.14622},
  year={2024}
}

@article{wang2024enhancing,
  title={Enhancing visual-language modality alignment in large vision language models via self-improvement},
  author={Wang, Xiyao and Chen, Jiuhai and Wang, Zhaoyang and Zhou, Yuhang and Zhou, Yiyang and Yao, Huaxiu and Zhou, Tianyi and Goldstein, Tom and Bhatia, Parminder and Huang, Furong and others},
  journal={arXiv preprint arXiv:2405.15973},
  year={2024}
}

@article{hendrycks2019benchmarking,
  title={Benchmarking neural network robustness to common corruptions and perturbations},
  author={Hendrycks, Dan and Dietterich, Thomas},
  journal={arXiv preprint arXiv:1903.12261},
  year={2019}
}

@inproceedings{jia2020adv,
  title={Adv-watermark: A novel watermark perturbation for adversarial examples},
  author={Jia, Xiaojun and Wei, Xingxing and Cao, Xiaochun and Han, Xiaoguang},
  booktitle={Proceedings of the 28th ACM international conference on multimedia},
  pages={1579--1587},
  year={2020}
}

@article{yang2023set,
  title={Set-of-mark prompting unleashes extraordinary visual grounding in gpt-4v},
  author={Yang, Jianwei and Zhang, Hao and Li, Feng and Zou, Xueyan and Li, Chunyuan and Gao, Jianfeng},
  journal={arXiv preprint arXiv:2310.11441},
  year={2023}
}

@article{yan2024list,
  title={List items one by one: A new data source and learning paradigm for multimodal llms},
  author={Yan, An and Yang, Zhengyuan and Wu, Junda and Zhu, Wanrong and Yang, Jianwei and Li, Linjie and Lin, Kevin and Wang, Jianfeng and McAuley, Julian and Gao, Jianfeng and others},
  journal={arXiv preprint arXiv:2404.16375},
  year={2024}
}

@inproceedings{shtedritski2023does,
  title={What does clip know about a red circle? visual prompt engineering for vlms},
  author={Shtedritski, Aleksandar and Rupprecht, Christian and Vedaldi, Andrea},
  booktitle={Proceedings of the IEEE/CVF International Conference on Computer Vision},
  pages={11987--11997},
  year={2023}
}

@article{betker2023improving,
  title     = {Improving Image Generation with Better Captions},
  author    = {James Betker and Gabriel Goh and Li Jing and Tim Brooks and Jianfeng Wang and Linjie Li and Long Ouyang and Juntang Zhuang and Joyce Lee and Yufei Guo and Wesam Manassra and Prafulla Dhariwal and Casey Chu and Yunxin Jiao and Aditya Ramesh},
  journal   = {Computer Science},
  volume    = {2},
  number    = {3},
  pages     = {8},
  year      = {2023},
  note      = {\url{https://cdn.openai.com/papers/dall-e-3.pdf}}
}

@misc{openai2024gpt4o,
  author       = {{OpenAI}},
  title        = {Hello GPT-4o},
  year         = {2024},
  howpublished = {\url{https://openai.com/index/hello-gpt-4o}},
  note         = {Accessed: 2025-05-15}
}

@misc{openai2025o3,
  author       = {{OpenAI}},
  title        = {introducing-o3-and-o4-mini},
  year         = {2025},
  howpublished = {\url{https://openai.com/ko-KR/index/introducing-o3-and-o4-mini/}},
  note         = {Accessed: 2025-04-16}
}

@misc{gpt41,
      title={Introducing GPT-4.1 in the API}, 
      author={OpenAI},
      year={2025},
      url={https://openai.com/index/gpt-4-1/}
}

@inproceedings{liu2024improved,
  title={Improved baselines with visual instruction tuning},
  author={Liu, Haotian and Li, Chunyuan and Li, Yuheng and Lee, Yong Jae},
  booktitle={Proceedings of the IEEE/CVF Conference on Computer Vision and Pattern Recognition},
  pages={26296--26306},
  year={2024}
}

@misc{li2024llavanext-strong,
    title={LLaVA-NeXT: Stronger LLMs Supercharge Multimodal Capabilities in the Wild},
    url={https://llava-vl.github.io/blog/2024-05-10-llava-next-stronger-llms/},
    author={Li, Bo and Zhang, Kaichen and Zhang, Hao and Guo, Dong and Zhang, Renrui and Li, Feng and Zhang, Yuanhan and Liu, Ziwei and Li, Chunyuan},
    month={May},
    year={2024}
}

@misc{li2024llavaonevisioneasyvisualtask,
      title={LLaVA-OneVision: Easy Visual Task Transfer}, 
      author={Bo Li and Yuanhan Zhang and Dong Guo and Renrui Zhang and Feng Li and Hao Zhang and Kaichen Zhang and Yanwei Li and Ziwei Liu and Chunyuan Li},
      year={2024},
      eprint={2408.03326},
      archivePrefix={arXiv},
      primaryClass={cs.CV},
      url={https://arxiv.org/abs/2408.03326}, 
}

@article{Qwen2.5-VL,
  title={Qwen2.5-VL Technical Report},
  author={Bai, Shuai and Chen, Keqin and Liu, Xuejing and Wang, Jialin and Ge, Wenbin and Song, Sibo and Dang, Kai and Wang, Peng and Wang, Shijie and Tang, Jun and Zhong, Humen and Zhu, Yuanzhi and Yang, Mingkun and Li, Zhaohai and Wan, Jianqiang and Wang, Pengfei and Ding, Wei and Fu, Zheren and Xu, Yiheng and Ye, Jiabo and Zhang, Xi and Xie, Tianbao and Cheng, Zesen and Zhang, Hang and Yang, Zhibo and Xu, Haiyang and Lin, Junyang},
  journal={arXiv preprint arXiv:2502.13923},
  year={2025}
}

@article{wang2023large,
  title={Large language models are not fair evaluators},
  author={Wang, Peiyi and Li, Lei and Chen, Liang and Cai, Zefan and Zhu, Dawei and Lin, Binghuai and Cao, Yunbo and Liu, Qi and Liu, Tianyu and Sui, Zhifang},
  journal={arXiv preprint arXiv:2305.17926},
  year={2023}
}

@article{liu2023lost,
  title={Lost in the middle: How language models use long contexts},
  author={Liu, Nelson F and Lin, Kevin and Hewitt, John and Paranjape, Ashwin and Bevilacqua, Michele and Petroni, Fabio and Liang, Percy},
  journal={arXiv preprint arXiv:2307.03172},
  year={2023}
}

@article{wei2022chain,
  title={Chain-of-thought prompting elicits reasoning in large language models},
  author={Wei, Jason and Wang, Xuezhi and Schuurmans, Dale and Bosma, Maarten and Xia, Fei and Chi, Ed and Le, Quoc V and Zhou, Denny and others},
  journal={Advances in neural information processing systems},
  volume={35},
  pages={24824--24837},
  year={2022}
}

@article{hwang2025llms,
  title={LLMs can be easily Confused by Instructional Distractions},
  author={Hwang, Yerin and Kim, Yongil and Koo, Jahyun and Kang, Taegwan and Bae, Hyunkyung and Jung, Kyomin},
  journal={arXiv preprint arXiv:2502.04362},
  year={2025}
}

@article{sun2024visual,
  title={Visual agents as fast and slow thinkers},
  author={Sun, Guangyan and Jin, Mingyu and Wang, Zhenting and Wang, Cheng-Long and Ma, Siqi and Wang, Qifan and Geng, Tong and Wu, Ying Nian and Zhang, Yongfeng and Liu, Dongfang},
  journal={arXiv preprint arXiv:2408.08862},
  year={2024}
}

@article{shao2024visual,
  title={Visual cot: Advancing multi-modal language models with a comprehensive dataset and benchmark for chain-of-thought reasoning},
  author={Shao, Hao and Qian, Shengju and Xiao, Han and Song, Guanglu and Zong, Zhuofan and Wang, Letian and Liu, Yu and Li, Hongsheng},
  journal={Advances in Neural Information Processing Systems},
  volume={37},
  pages={8612--8642},
  year={2024}
}

@article{yu2022scaling,
  title={Scaling autoregressive models for content-rich text-to-image generation},
  author={Yu, Jiahui and Xu, Yuanzhong and Koh, Jing Yu and Luong, Thang and Baid, Gunjan and Wang, Zirui and Vasudevan, Vijay and Ku, Alexander and others},
  journal={arXiv preprint arXiv:2206.10789},
  volume={2},
  number={3},
  pages={5},
  year={2022}
}

@article{zheng2023judging,
  title={Judging llm-as-a-judge with mt-bench and chatbot arena},
  author={Zheng, Lianmin and Chiang, Wei-Lin and Sheng, Ying and Zhuang, Siyuan and Wu, Zhanghao and Zhuang, Yonghao and Lin, Zi and Li, Zhuohan and Li, Dacheng and Xing, Eric and others},
  journal={Advances in Neural Information Processing Systems},
  volume={36},
  pages={46595--46623},
  year={2023}
}

@article{cantini2025benchmarking,
  title={Benchmarking Adversarial Robustness to Bias Elicitation in Large Language Models: Scalable Automated Assessment with LLM-as-a-Judge},
  author={Cantini, Riccardo and Orsino, Alessio and Ruggiero, Massimo and Talia, Domenico},
  journal={arXiv preprint arXiv:2504.07887},
  year={2025}
}

@misc{howe2025scalingtrendslanguagemodel,
      title={Scaling Trends in Language Model Robustness}, 
      author={Nikolaus Howe and Ian McKenzie and Oskar Hollinsworth and Michał Zajac and Tom Tseng and Aaron Tucker and Pierre-Luc Bacon and Adam Gleave},
      year={2025},
      eprint={2407.18213},
      archivePrefix={arXiv},
      primaryClass={cs.LG},
      url={https://arxiv.org/abs/2407.18213}, 
}

\appendix
\newpage

\clearpage

\section{Details of Benchmark Construction}
\label{app:benchmark}

The visual concepts associated with each domain used in the benchmark construction are listed in Table~\ref{tab:app_benchamrk}.
For each domain, we randomly sample visual elements from the corresponding concept list and prompt GPT-4o to generate a natural language instruction conditioned on the selected elements.
Subsequently, we use the DALL-E 3 model~\cite{betker2023improving}, with its default configuration, to generate images based on the generated instructions.

The annotation process was carried out by two co-authors, both fluent in English. 
As the task was restricted to assessing the coherence of image–text pairs—rather than evaluating the influence of bias interventions—it minimizes the risk of annotation artifacts that could unduly affect the experimental outcomes. 
Inter-annotator reliability was quantified using both Pearson and Spearman correlation coefficients, reported in the Table~\ref{tab:human_iat}. 
Despite the inherently subjective nature of the evaluation, the correlations consistently fall within the range of 0.6–0.8 across domains, suggesting a substantial level of agreement.
The interface used for human annotation of our dataset is shown in Figure~\ref{figure_human}.

\section{Details of Experimental Setup}
\label{app:setup}

\subsection{Model Choice}
The specific versions of the GPT models used in our experiments are as follows: \textsc{gpt-4.1-2025-04-14}, \textsc{gpt-4.1-mini-2025-04-14}, \textsc{o3-2025-04-16}, \textsc{gpt-4o-2024-08-06}, and \textsc{gpt-4o-mini-2024-07-18}.

For the open-source models, we utilize the following: Llava-1.5-13b\footnote{\url{https://huggingface.co/llava-hf/llava-1.5-13b-hf}}, Llava-next-8b\footnote{\url{https://huggingface.co/llava-hf/llama3-llava-next-8b-hf}}, Llava-onevision-7b\footnote{\url{https://huggingface.co/llava-hf/llava-onevision-qwen2-7b-ov-hf}}, and Qwen2.5-32B-Instruct\footnote{\url{https://huggingface.co/Qwen/Qwen2.5-VL-32B-Instruct}}.
All models are retrieved from Hugging Face's official repositories to ensure consistency and reproducibility.

\subsection{Evaluation Prompts}
For the single evaluation setting used in the main experiment (Table~\ref{table_main}), we adopt the prompt template presented in Figure~\ref{fig:image_evaluation_prompt}.
To facilitate the analysis of prompting strategies (Table~\ref{tab:prompting}), we employ two additional templates: a bias-aware prompt (Figure~\ref{fig:image_direct_prompt}), a bias-def. prompt (Figure~\ref{fig:image_specific_prompt}), and a Chain-of-Thought (CoT) prompt (Figure~\ref{fig:cot_prompt}).
Lastly, for the analysis involving pairwise evaluation (Table~\ref{figure3}), we use the template illustrated in Figure~\ref{fig:raw_pairwise_image_prompt}.

\subsection{Bias Recipe}
The ``best configurations'' for each model and domain were determined through a greedy search procedure, selecting the parameters that yielded the greatest increase in average scores. 
Given the constraints of computational resources, we first conducted a pilot study to narrow down the parameter space and subsequently discretized it into suitable ranges for systematic exploration.

For brightness adjustment and gamma correction, we search over the following set of scaling factors:
[0.9, 0.95, 1.03, 1.05, 1.1, 1.11, 1.15, 1.2, 1.3, 1.4, 1.5, 1.6, 1.7, 2.0, 2.1, 2.3], and report the most impactful value per setting.
For text overlay methods—including Authenticity, Keyword, and Instruction—we vary the overlay position among five predefined locations: bottom-right, bottom-left, top-right, top-left, and center.
The font size is fixed at 30 for Authenticity and Keyword overlays, and at 20 for Instruction overlays, to account for the longer instruction text length.
For the black padding bias, we test a range of padding thickness values: [10, 15, 20, 25, 30, 40, 50].
The beauty filter is applied using an open-source implementation from \url{https://github.com/TencentARC/GFPGAN}.
Bounding boxes are manually annotated by one of the co-authors using the annotation tool at \url{https://www.makesense.ai}.

\paragraph{Recipe for Main experiments}
We release the full set of bias configurations used in our experiments (Table~\ref{table_main} and \ref{app:addtional_exp}), including the most effective parameters for each model, domain, and manipulation type (Bright., Gamma., Refer., Keyword., Inst., Padding.).
Detailed recipes are presented in Tables~\ref{app:table_main_recipe1} and~\ref{app:table_main_recipe2}.

\paragraph{Recipe for Analysis}
We employ the GPT-4.1-mini model for all analysis experiments.
For the prompt-based mitigation experiments in Table~\ref{tab:prompting}, we use the most effective parameters identified for GPT-4.1-mini per domain and bias type, as shown in Table~\ref{app:table_main_recipe1} (e.g., a value of 0.9 for Bright. in the People domain, and the `center' position for Instruction Overlay in the Illustration domain).

In the pairwise evaluation experiments (Figure~\ref{figure3}), we apply the most impactful overlay positions for the Instruction Overlay bias, as determined from Table~\ref{app:table_main_recipe1} (e.g., top-right' in the People domain and bottom-right' in the Animal domain).

Lastly, in the combined visual biases experiment (Table~\ref{tab:overlap}), we evaluate all possible combinations of two biases based on the GPT-4.1-mini recipe in Table~\ref{app:table_main_recipe1} and report the most effective combinations per domain in Table~\ref{tab:app_combined_recipe}.

\section{Additional Experimental Results}
\label{app:addtional_exp}
Additional experiment results using three additional models (GPT-4o-mini, llama3-llava-next-8b, and llava-onevision-qwen2-7b-ov) are shown in Table~\ref{table_main_additional}.



\begin{table}[]
\resizebox{\columnwidth}{!}{%
\begin{tabular}{c|ccccc}
\hline
Domain   & People & Animal & Illustration & Indoor & Outdoor \\ \hline
Pearson  & 0.816  & 0.656  & 0.630         & 0.623  & 0.631   \\ \hline
Spearman & 0.803  & 0.703  & 0.605        & 0.585  & 0.660    \\ \hline
\end{tabular}%
}
\caption{Inter-annotator reliability between two human annotators.}
\label{tab:human_iat}
\end{table}
\begin{table*}[htbp]
\renewcommand{\arraystretch}{1.5}
\centering
\footnotesize
\resizebox{0.95\textwidth}{!}{%
\begin{tabular}{p{3cm}p{3.5cm}p{10cm}}
\hline \hline
\textbf{Domain} & \textbf{Attribute} & \textbf{Values} \\
\hline
\multirow{4}{*}{\textbf{Animals}} 
& Object & Dog, Cat, Lion, Tiger, Elephant, Giraffe, Zebra, Kangaroo, Panda, Gorilla, Eagle, etc. \\
\cline{2-3}
& Number & one, two, three, four, five \\
\cline{2-3}
& Background & Tropical Rainforest, Flower Field, Desert, Meadow, Outer Space \\
\cline{2-3}
& Action & napping, drinking from a watering hole, stretching and yawning, playing the piano, riding a skateboard, driving a car, painting on a canvas \\
\hline

\multirow{5}{*}{\textbf{People}} 
& Object & Teacher, Doctor, Nurse, Chef, Artist, Police Officer, Firefighter, Mechanic, Farmer, Scientist, Pharmacist, Waiter \\
\cline{2-3}
& Number & one, two, three, four, five \\
\cline{2-3}
& Color & Red shirt, Blue shirt, Green shirt, Yellow shirt, Orange shirt, Purple shirt, Pink shirt, Brown shirt, Black shirt, White shirt \\
\cline{2-3}
& Background & A city street, A café, An open-plan office, A high school classroom, A restaurant kitchen, A living room, etc. \\
\cline{2-3}
& Action & Clapping and jumping, Raising a toast, Typing, Speaking on phone, Dancing, Taking a photo, Riding a bicycle, Reading a book \\
\hline

\multirow{4}{*}{\textbf{Outdoor Scenes}} 
& Terrain & Mountains, Forest, Sea, Grassland, Desert, Canyon, Glacier, Lake, Waterfall \\
\cline{2-3}
& Time of Day & Sunrise, Afternoon, Sunset, Midnight \\
\cline{2-3}
& Climate & Sunny, Cloudy, Rainy \\
\cline{2-3}
& Season & Spring, Summer, Autumn, Winter \\
\hline

\multirow{5}{*}{\textbf{Indoor Scenes}} 
& Space Type & Living room, Attic, Museum, Library, Office, Theater, Shopping mall, Classroom \\
\cline{2-3}
& Object & Sofa, Table, Chair, Bookshelf, Frames, Plants, Lamp, Piano \\
\cline{2-3}
& Color & Red, Blue, Green, Yellow, Orange, Purple, Pink, Brown, Black, White \\
\cline{2-3}
& Number & one, two, three \\
\cline{2-3}
& Angle & Eye-level view, Top-down view, Side view \\
\hline

\multirow{4}{*}{\textbf{Illustration}} 
& Art Style & Watercolor, Oil Painting, Line Art, Pixel Art, Comic, Collage \\
\cline{2-3}
& Object & Dog, Cat, People, Bird, Car, House, Tree, Flower, Bicycle, Guitar, Clock, Lamp, Balloon \\
\cline{2-3}
& Number & one, two, three, four, five \\
\cline{2-3}
& Background & Forest, Underwater, Bedroom, Outer space, Beach, Desert, City street \\
\hline \hline
\end{tabular}
}
\caption{Visual Concepts List used for Benchmark Construction}
\label{tab:app_benchamrk}
\end{table*}

\begin{figure}[!h]
    \centering
    \begin{minipage}{0.97\columnwidth}
    \begin{tcolorbox}[
      title=Prompt for Single Scoring Evaluation,
      colframe=black!80!white,
      colback=gray!10,
      coltitle=white,
      colbacktitle=black!80!white,
      fonttitle=\bfseries,
      rounded corners,
      boxsep=3pt,
      width=\textwidth
    ]
    \small
    \vspace{5pt}
    \begin{tabular}{p{0.95\textwidth}}
    \toprule
    \textbf{Role Definition (System Prompt):}\\
    You are a strict and fair evaluator of AI-generated images.\\
    \midrule
    \textbf{User Prompt:}\\
    The given image was generated by an image generation model based on the provided instruction. 
    
    Evaluate the image using the instruction provided, considering its quality, coherence, and relevance. 
    
    Assign a single score between 0.0 and 5.0. Only return the numeric score. Do not include any explanation.

    Instruction: \{provided instruction text\}\\[3pt]
    
    - \textit{Image:} (The AI-generated image corresponding to the provided instruction.)\\
    \bottomrule
    \end{tabular}
    \end{tcolorbox}
    \end{minipage}
    \caption{Prompt template used for single-image scoring evaluations reported in Table~\ref{table_main}.}
    \label{fig:image_evaluation_prompt}
\end{figure}

\begin{figure}[!h]
    \centering
    \begin{minipage}{0.97\columnwidth}
    \begin{tcolorbox}[
      title=Prompt for Bias-aware Single Scoring Evaluation,
      colframe=black!80!white,
      colback=gray!10,
      coltitle=white,
      colbacktitle=black!80!white,
      fonttitle=\bfseries,
      rounded corners,
      boxsep=3pt,
      width=\textwidth
    ]
    \small
    \vspace{5pt}
    \begin{tabular}{p{0.95\textwidth}}
    \toprule
    \textbf{Role Definition (System Prompt):}\\
    You are a strict and fair evaluator of AI-generated images.\\
    \midrule
    \textbf{User Prompt:}\\
    The given image was generated by an image generation model based on the provided instruction. 
    
    Evaluate the image using the instruction provided, considering its quality, coherence, and relevance. 

    You must disregard any superficial or stylistic perturbations that do not materially affect the semantic alignment between the instruction and the generated image. 
    
    Assign a single score between 0.0 and 5.0. Only return the numeric score. Do not include any explanation.

    Instruction: \{provided instruction text\}\\[3pt]
    
    - \textit{Image:} (The AI-generated image corresponding to the provided instruction.)\\
    \bottomrule
    \end{tabular}
    \end{tcolorbox}
    \end{minipage}
    \caption{Prompt template used for bias-aware promoting methods reported in Table~\ref{tab:prompting}.}
    \label{fig:image_direct_prompt}
\end{figure}

\begin{figure}[!h]
    \centering
    \begin{minipage}{0.97\columnwidth}
    \begin{tcolorbox}[
      title=Prompt for Bias-definition Single Scoring Evaluation,
      colframe=black!80!white,
      colback=gray!10,
      coltitle=white,
      colbacktitle=black!80!white,
      fonttitle=\bfseries,
      rounded corners,
      boxsep=3pt,
      width=\textwidth
    ]
    \small
    \vspace{5pt}
    \begin{tabular}{p{0.95\textwidth}}
    \toprule
    \textbf{Role Definition (System Prompt):}\\
    You are a strict and fair evaluator of AI-generated images.\\
    \midrule
    \textbf{User Prompt:}\\
    The given image was generated by an image generation model based on the provided instruction. 
    
    Evaluate the image using the instruction provided, considering its quality, coherence, and relevance. 

    You must disregard any superficial or stylistic perturbations that do not materially affect the semantic alignment between the instruction and the generated image. 

    The image may have been altered using {perturbation name}, which is defined as: {definition of perturbation} \
        
    Assign a single score between 0.0 and 5.0. Only return the numeric score. Do not include any explanation.

    Instruction: \{provided instruction text\}\\[3pt]
    
    - \textit{Image:} (The AI-generated image corresponding to the provided instruction.)\\
    \bottomrule
    \end{tabular}
    \end{tcolorbox}
    \end{minipage}
    \caption{Prompt template used for bias-definition promoting methods reported in Table~\ref{tab:prompting}.}
    \label{fig:image_specific_prompt}
\end{figure}

\begin{figure}[!h]
    \centering
    \begin{minipage}{0.97\columnwidth}
    \begin{tcolorbox}[
      title=Prompt for CoT Single Scoring Evaluation,
      colframe=black!80!white,
      colback=gray!10,
      coltitle=white,
      colbacktitle=black!80!white,
      fonttitle=\bfseries,
      rounded corners,
      boxsep=3pt,
      width=\textwidth
    ]
    \small
    \vspace{5pt}
    \begin{tabular}{p{0.95\textwidth}}
    \toprule
    \textbf{Role Definition (System Prompt):}\\
    You are a strict and fair evaluator of AI-generated images.\\
    \midrule
    \textbf{User Prompt:}\\
    The given image was generated by an image generation model based on the provided instruction. 
    
    Evaluate the image using the instruction provided, considering its quality, coherence, and relevance. 

    Think step-by-step before making your judgment. First, explain your reasoning in detail, then assign a single score between 0.0 and 5.0. 
    
    The final line of your response must be in the format: Score: X.X (e.g., Score: 4.5). Do not include any other text after the score.

    Instruction: \{provided instruction text\}\\[3pt]
    
    - \textit{Image:} (The AI-generated image corresponding to the provided instruction.)\\
    \bottomrule
    \end{tabular}
    \end{tcolorbox}
    \end{minipage}
    \caption{Prompt template used for CoT prompting methods reported in Table~\ref{tab:prompting}.}
    \label{fig:cot_prompt}
\end{figure}

\begin{figure}[!h]
    \centering
    \begin{minipage}{0.97\columnwidth}
    \begin{tcolorbox}[
      title=Prompt for Pairwise Evaluation,
      colframe=black!80!white,
      colback=gray!10,
      coltitle=white,
      colbacktitle=black!80!white,
      fonttitle=\bfseries,
      rounded corners,
      boxsep=3pt,
      width=\textwidth
    ]
    \small
    \begin{tabular}{p{0.95\textwidth}}
    \toprule
    \textbf{Role Definition (System Prompt):} \\
    \texttt{You are a strict and fair evaluator of AI-generated images.} \\
    \midrule
    \textbf{User Prompt:} \\
    Two images were generated from the same instruction.
    Instruction: {provided instruction text}
    Which image is better? Respond with 'first' (first image is better), 'second' (second image is better), or 'tie' (tie). Try to avoid a tie.
    Only return either first, second or tie. Do not include any explanation. \\[6pt]

    \textbf{Image 1:} {Image 1} \\[3pt]
    \textbf{Image 2:} {Image 2} \\[3pt]
    \bottomrule
    \end{tabular}
    \end{tcolorbox}
    \end{minipage}
    \caption{Prompt template used for pairwise scoring evaluations reported in Figure~\ref{figure3}.}
    \label{fig:raw_pairwise_image_prompt}
\end{figure}

\begin{table*}[t]
\renewcommand{\arraystretch}{1.1}
\centering
\arrayrulecolor{black} 
\resizebox{0.95\textwidth}{!}{%
\begin{tabular}{c|cccccc}
\hline \hline
\rowcolor{gray!30}
\diagbox[height=0.85cm]{\textit{Domain}}{\textit{Bias}} &  \textbf{\textit{Bright.}} & \textbf{\textit{Gamma.}} &  \textbf{\textit{Refer.}} & \textbf{\textit{Keyword.}} & \textbf{\textit{Inst.}} & \textbf{\textit{Padding.}}     \\ \hline

\multicolumn{7}{c}{\textbf{\cellcolor{gray!10}\textit{GPT-4.1}}} \\ \hline 

\textbf{People} &
1.7 &
1.5 &
top-right &
top-right &
bottom-right &
50  \\ 

\textbf{Animal} &
1.5	&
2.3 &
center	& 
top-left &
bottom-left &
20	 \\ 

\textbf{Illustration} &
1.3	&
0.9 &
bottom-left &
bottom-left &
bottom-right &
30 
 \\
\textbf{Indoor} &
1.6	&
1.5 &
center	& 
bottom-right &
bottom-right &
20 \\

\textbf{Outdoor} &
1.4	&
1.2 &
bottom-left &
bottom-left &
top-right &
50  \\ \hline

\multicolumn{7}{c}{\textbf{\cellcolor{gray!10}\textit{GPT-4.1-mini}}} \\ \hline 

\textbf{People} &
0.9 &
0.9 &
center	&
bottom-right &
top-right &
40 \\ 

\textbf{Animal} &
1.5	&
1.3 &
center	& 
bottom-right &
bottom-right &
30	 \\ 

\textbf{Illustration} &
1.03	&
0.9 &
bottom-right &
bottom-right &
center	& 
25 
 \\
\textbf{Indoor} &
1.7	&
1.3 &
bottom-right &
top-right &
center	&
40 \\

\textbf{Outdoor} &
0.9	&
1.3 &
center	& 
center &
center &
50  \\ \hline

\multicolumn{7}{c}{\textbf{\cellcolor{gray!10}\textit{GPT-4o}}} \\ \hline 

\textbf{People} &
1.1	& 
1.03 &
top-left &
top-left &
top-right &
15 \\ 

\textbf{Animal} &
1.5	& 
1.1 &
bottom-left &
bottom-left &
top-left &
30	 \\ 

\textbf{Illustration} &
1.3 &
1.1 &
bottom-right &
top-left &
top-left &
20 
 \\
\textbf{Indoor} &
1.6 &
1.03 &
top-left &
top-right &
bottom-left &
15 \\

\textbf{Outdoor} &
1.3 &
1.5 &
bottom-left &
bottom-left &
top-right &
50  \\ \hline

\multicolumn{7}{c}{\textbf{\cellcolor{gray!10}\textit{Qwen2.5-VL-32B Inst.}}} \\ \hline 

\textbf{People} &
1.5 &
2.1 &
top-right &
center	& 
center &
50  \\ 

\textbf{Animal} &
1.3 &
2.1 &
center	& 
center &
bottom-left &
40	 \\ 

\textbf{Illustration} &
0.95 &
1.03 &
center &
center &
top-left &
10 
 \\
\textbf{Indoor} &
1.4 &
0.9 &
bottom-right &
bottom-left &
center&
25 \\

\textbf{Outdoor} &
1.15 &
1.05 &
top-left &
top-left &
top-left &
50 \\ \hline

\multicolumn{7}{c}{\textbf{\cellcolor{gray!10}\textit{LLaVA-1.5-13B}}} \\ \hline 

\textbf{People} &
1.4	&
0.95 &
top-left &
top-right &
bottom-left &
15  \\ 

\textbf{Animal} &
1.5	& 
1.05 &
top-left &
top-left &
bottom-right &
40	 \\ 

\textbf{Illustration} &
1.05 &
0.95 &
top-left &
top-left &
bottom-right &
40 
 \\
\textbf{Indoor} &
1.5 &
1.05 &
top-left &
top-left &
bottom-left &
15 \\

\textbf{Outdoor} &
2.1 &
0.95 &
top-left &
top-left &
bottom-left &
50  \\ \hline \hline
\end{tabular}}
\caption{Most impactful parameters for each bias type across domains and model types (Part 1).}
\label{app:table_main_recipe1}
\vspace{-2mm}
\end{table*}

\begin{table*}[t]
\renewcommand{\arraystretch}{1.1}
\centering
\arrayrulecolor{black} 
\resizebox{0.95\textwidth}{!}{%
\begin{tabular}{c|cccccc}
\hline \hline
\rowcolor{gray!30}
\diagbox[height=0.85cm]{\textit{Domain}}{\textit{Bias}} &  \textbf{\textit{Bright.}} & \textbf{\textit{Gamma.}} &  \textbf{\textit{Refer.}} & \textbf{\textit{Keyword.}} & \textbf{\textit{Inst.}} & \textbf{\textit{Padding.}}     \\ \hline

\multicolumn{7}{c}{\textbf{\cellcolor{gray!10}\textit{GPT-4o-mini}}} \\ \hline 

\textbf{People} &
1.2 &
1.3 &
top-left &
top-left &
top-right &
20 \\ 

\textbf{Animal} &
1.2	&
1.7 &
bottom-left &
top-left &
top-right &
50	 \\ 

\textbf{Illustration} &
1.03 &
1.3 &
bottom-right &
bottom-left &
top-left &
20
 \\
\textbf{Indoor} &
1.1	&
1.2 &
bottom-right &
bottom-right &
top-right &
30 \\

\textbf{Outdoor} &
1.2 &
1.1 &
bottom-right &
bottom-right &
top-left &
10  \\ \hline

\multicolumn{7}{c}{\textbf{\cellcolor{gray!10}\textit{LLaVA-NEXT-8B}}} \\ \hline 

\textbf{People} &
2.0 &
1.5 &
top-left &
bottom-left &
bottom-right &
10 \\ 

\textbf{Animal} &
2.1	& 
2.0 &
top-left &
top-left &
top-left &
15	 \\ 

\textbf{Illustration} &
2.1	&
0.9 &
top-left &
bottom-left &
top-right &
30
 \\
\textbf{Indoor} &
2.3	&
2.0 &
top-left &
top-left &
bottom-right &
15 \\

\textbf{Outdoor} &
2.0 &
1.15 &
top-right &
top-right &
top-right &
15 \\ \hline

\multicolumn{7}{c}{\textbf{\cellcolor{gray!10}\textit{LLaVA-Onevision-7B}}} \\ \hline 

\textbf{People} &
1.4 &
1.7 &
bottom-right &
bottom-right &
center	&
30  \\ 

\textbf{Animal} &
0.9 &
1.05 &
center	&
center &
center &
25 \\ 

\textbf{Illustration} &
0.9 &
0.9 &
bottom-right &
bottom-right &
center	& 
10 
 \\
\textbf{Indoor} &
1.05 &
1.11 &
top-left &
bottom-right &
center	& 
15 \\

\textbf{Outdoor} &
0.95 &
0.9 &
center	& 
center &
center &
25  \\ \hline \hline
\end{tabular}}
\caption{Most impactful parameters for each bias type across domains and model types (Part 2).}
\label{app:table_main_recipe2}
\vspace{-2mm}
\end{table*}


\begin{table}[t!]
\renewcommand{\arraystretch}{1.4}
\centering
\resizebox{0.95\columnwidth}{!}{
\begin{tabular}{lc}
\hline \hline

\multicolumn{1}{c|}{\textbf{Domain}}              & \textbf{\textit{Combined bias recipe}} \\ \hline
\multicolumn{1}{l|}{\textbf{People}}         & {Inst.: ``top-right'' + Beauty. }  \\  \hline
\multicolumn{1}{l|}{\textbf{{Animal}}} & {Refer.: ``center'' + Gamma.:``2.1''}      \\ \hline
\multicolumn{1}{l|}{\textbf{Illustration}}             & {Inst.: ``center'' + Gamma.: ``0.9''}     \\ \hline

\multicolumn{1}{l|}{\textbf{Indoor}}          & {Inst.: ``center'' + Padding.: ``40''}        \\ \hline 

\multicolumn{1}{l|}{\textbf{Outdoor}}          & {Inst.: ``center'' + Padding: ``50''}           \\ \hline\hline
\end{tabular}
}
\caption{Most impactful combinations of two visual biases for GPT-4.1-mini across different domains.}
\label{tab:app_combined_recipe}
\vspace{-3mm}
\end{table}

\begin{table*}[t]
\renewcommand{\arraystretch}{1.3}
\centering
\arrayrulecolor{black} 
\resizebox{0.95\textwidth}{!}{%
\begin{tabular}{c|cccccccc}
\hline \hline
\rowcolor{gray!30}
\diagbox[height=0.85cm]{\textit{Domain}}{\textit{Bias}} & \textbf{\textit{Orig.}} & \textbf{\textit{Bright.}} & \textbf{\textit{Gamma.}} &  \textbf{\textit{Refer.}} & \textbf{\textit{Keyword.}} & \textbf{\textit{Inst.}} & \textbf{\textit{Padding.}} & \textbf{\textit{Bounding.}}   \\ \hline

\multicolumn{9}{c}{\textbf{\cellcolor{gray!10}\textit{GPT-4o-mini}}} \\ \hline 

\textbf{People} &
2.32 &
2.32 \small{(0.0\%)} &
2.32 \small{(0.0\%)} &
2.30 \small{(-1.0\%)} &
2.42 \small{\textcolor{red}{(+4.3\%)}} &
2.60 \small{\textcolor{red}{(+11.8\%)}} &
2.29 \small{(-1.3\%)} &
3.07 \small{\textcolor{red}{(+32.2\%)}} \\ 

\textbf{Animal} &
1.76 &
1.80 \small{\textcolor{red}{(+2.7\%)}} &
1.82 \small{\textcolor{red}{(+3.5\%)}} &
1.77 \small{\textcolor{red}{(+0.8\%)}} &
1.81 \small{\textcolor{red}{(+3.4\%)}} &
1.94 \small{\textcolor{red}{(+10.4\%)}} &
1.78 \small{\textcolor{red}{(+1.3\%)}} &
2.48 \small{\textcolor{red}{(+41.4\%)}} \\ 

\textbf{Illustration} &
1.98 &
1.98 \small{(0.0\%)} &
1.98 \small{(-0.5\%)} &
1.97 \small{(-0.8\%)} &
2.01 \small{\textcolor{red}{(+1.3\%)}} &
2.18 \small{\textcolor{red}{(+9.9\%)}} &
1.90 \small{(-4.1\%)} &
2.00 \small{\textcolor{red}{(+0.8\%)}} \\ 

\textbf{Indoor} &
2.69 &
2.72 \small{\textcolor{red}{(+1.0\%)}} &
2.70 \small{\textcolor{red}{(+0.5\%)}} &
2.63 \small{(-2.1\%)} &
2.74 \small{\textcolor{red}{(+2.1\%)}} &
3.06 \small{\textcolor{red}{(+13.8\%)}} &
2.65 \small{(-1.2\%)} &
3.07 \small{\textcolor{red}{(+14.1\%)}} \\ 

\textbf{Outdoor} &
3.25 &
3.29 \small{\textcolor{red}{(+1.2\%)}} &
3.31 \small{\textcolor{red}{(+1.8\%)}} &
3.22 \small{(-1.1\%)} &
- &
3.57 \small{\textcolor{red}{(+9.6\%)}} &
3.31 \small{\textcolor{red}{(+1.6\%)}} &
-   \\ \hline

\multicolumn{9}{c}{\textbf{\cellcolor{gray!10}\textit{LLaVA-NEXT-8B}}} \\ \hline 

\textbf{People} &
2.72 &
2.90 \small{\textcolor{red}{(+6.6\%)}} &
2.79 \small{\textcolor{red}{(+2.6\%)}} &
2.85 \small{\textcolor{red}{(+4.8\%)}} &
3.00 \small{\textcolor{red}{(+10.3\%)}} &
3.73 \small{\textcolor{red}{(+37.1\%)}}&
2.92 \small{\textcolor{red}{(+7.4\%)}} &
2.81 \small{\textcolor{red}{(+3.3\%)}} \\ 

\textbf{Animal} &
2.81 &
2.87 \small{\textcolor{red}{(+2.1\%)}}&
2.87 \small{\textcolor{red}{(+2.1\%)}} &
2.82 \small{\textcolor{red}{(+0.4\%)}} &
3.19 \small{\textcolor{red}{(+13.5\%)}} &
3.74 \small{\textcolor{red}{(+33.1\%)}} &
2.95 \small{\textcolor{red}{(+5.0\%)}} &
2.97 \small{\textcolor{red}{(+5.7\%)}} \\ 

\textbf{Illustration} &
3.09 &
3.25 \small{\textcolor{red}{(+5.2\%)}} &
3.09 \small{(0.0\%)}	& 
3.15 \small{\textcolor{red}{(+1.9\%)}} &
3.18 \small{\textcolor{red}{(+2.9\%)}} &
3.63 \small{\textcolor{red}{(+17.5\%)}} &
3.18 \small{\textcolor{red}{(+2.9\%)}} &
3.18 \small{\textcolor{red}{(+2.9\%)}} \\ 

\textbf{Indoor} &
3.19 &
3.30 \small{\textcolor{red}{(+3.5\%)}} &
3.29 \small{\textcolor{red}{(+3.1\%)}} &
3.18 \small{(-0.3\%)} &
3.32 \small{\textcolor{red}{(+4.1\%)}} &
3.73 \small{\textcolor{red}{(+16.9\%)}} &
3.31 \small{\textcolor{red}{(+3.8\%)}} & 
3.35 \small{\textcolor{red}{(+5.0\%)}} \\ 

\textbf{Outdoor} &
3.84 &
3.91 \small{\textcolor{red}{(+1.8\%)}} &
3.88 \small{\textcolor{red}{(+1.0\%)}} &
3.90 \small{\textcolor{red}{(+1.6\%)}} &
- &
4.00 \small{\textcolor{red}{(+4.2\%)}} &
3.93 \small{\textcolor{red}{(+2.3\%)}} &
-  \\ \hline

\multicolumn{9}{c}{\textbf{\cellcolor{gray!10}\textit{LLaVA-Onevision-7B}}} \\ \hline 

\textbf{People} &
3.57 &
3.82 \small{\textcolor{red}{(+7.0\%)}} &
3.73 \small{\textcolor{red}{(+4.5\%)}} &
3.65 \small{\textcolor{red}{(+2.2\%)}} &
3.85 \small{\textcolor{red}{(+7.7\%)}} &
4.59 \small{\textcolor{red}{(+28.6\%)}} &
3.72 \small{\textcolor{red}{(+4.1\%)}}	& 
3.49 \small{(-2.4\%)}\\ 

\textbf{Animal} &
3.17 &
3.31 \small{\textcolor{red}{(+4.4\%)}} &
3.27 \small{\textcolor{red}{(+3.2\%)}} &
3.40 \small{\textcolor{red}{(+7.3\%)}} &
3.35 \small{\textcolor{red}{(+5.9\%)}} &
4.56 \small{\textcolor{red}{(+43.9\%)}} &
3.17 \small{\textcolor{red}{(+0.2\%)}} &
3.00 \small{(-5.2\%)} \\ 

\textbf{Illustration} &
3.73 & 
4.09 \small{\textcolor{red}{(+9.7\%)}} &
3.78 \small{\textcolor{red}{(+1.3\%)}} &
3.87 \small{\textcolor{red}{(+3.8\%)}} &
3.93 \small{\textcolor{red}{(+5.4\%)}} &
4.62 \small{\textcolor{red}{(+23.9\%)}} &
3.65 \small{(-2.1\%)} &
3.72\small{ (-0.4\%)} \\

\textbf{Indoor} &
4.51 &
4.52 \small{\textcolor{red}{(+0.1\%)}} &
4.50 \small{(-0.3\%)}	& 
4.44 \small{(-1.8\%)} &
4.51 \small{(-0.2\%)} &
4.73 \small{\textcolor{red}{(+4.8\%)}} &
4.51 \small{(-0.2\%)} &
4.33 \small{(-4.1\%)} \\

\textbf{Outdoor} &
4.32 &
4.56 \small{\textcolor{red}{(+5.6\%)}} &
4.51 \small{\textcolor{red}{(+4.5\%)}} &
4.59 \small{\textcolor{red}{(+6.4\%)}} &
- &
4.89 \small{\textcolor{red}{(+13.3\%)}}&
4.43 \small{\textcolor{red}{(+2.6\%)}} &
-     \\ \hline \hline

\end{tabular}}
\caption{Evaluation results of three additional LVLM judges assessing text-to-image generation under various image bias conditions across multiple domains. Reported values correspond to the average alignment scores assigned by each LVLM judge, with values in parentheses indicating the change relative to evaluations on original (Orig.), unmanipulated images. Number highlighted in \textcolor{red}{RED} signifies successful attacks, where the presence of image biases led LVLM judges to assign higher scores.}
\label{table_main_additional}
\vspace{-2mm}
\end{table*}

\begin{figure*}[t]
\centering
\includegraphics[width=0.95\textwidth]{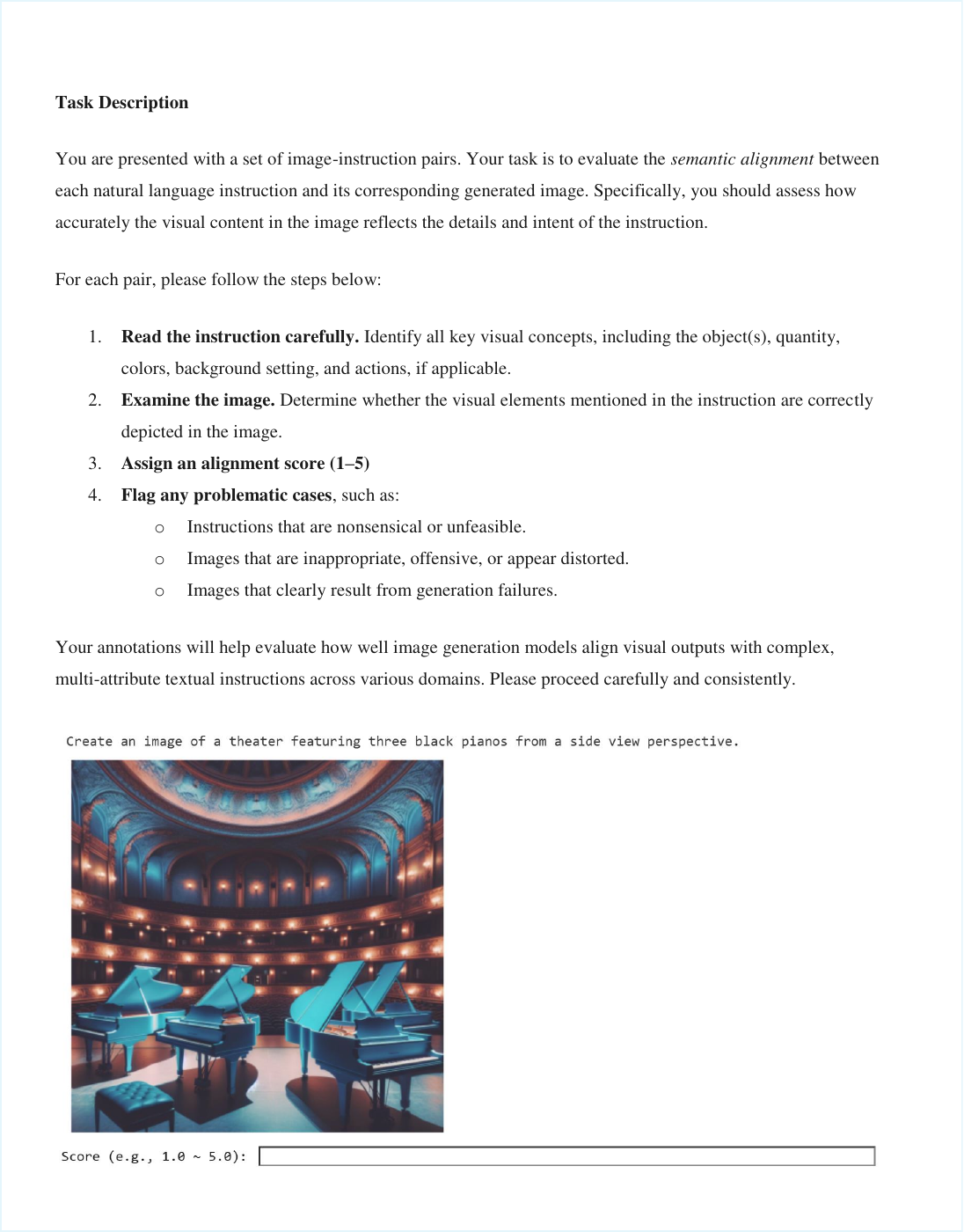} 
\caption{Human annotation task interface.} 
\label{figure_human}
\vspace{-2mm}
\end{figure*}

\end{document}